\documentclass{article}
\usepackage{hyperref}
\pdfoutput=1
\PassOptionsToPackage{numbers, compress}{natbib}

\usepackage[preprint]{nips_2018}

\usepackage[utf8]{inputenc} 
\usepackage[T1]{fontenc}    
\usepackage{url}            
\usepackage{booktabs}       
\usepackage{amsfonts}       
\usepackage{nicefrac}       
\usepackage{microtype}      
\usepackage{graphicx,subfigure}
\graphicspath{{./figures/}}
\usepackage{color}
\usepackage[toc,page]{appendix}
\usepackage{multirow}
\usepackage{arydshln}
\usepackage{enumitem}
\title{The Unreasonable Effectiveness of Texture Transfer for Single Image Super-resolution}

\author{
  Muhammad Waleed Gondal \\
  Max Planck Institute for Intelligent Systems.\\
  \texttt{waleed.gondal@tuebingen.mpg.de} \\
  \And
  Bernhard Schölkopf \\
  Max Planck Institute for Intelligent Systems.\\
  \texttt{bs@tuebingen.mpg.de} \\
  \AND
  Michael Hirsch \\
  Amazon Research.\\
  \texttt{hirsch@amazon.com} \\
}

\begin{document}

\maketitle

\begin{abstract}
While implicit generative models such as GANs have shown impressive results in high quality image reconstruction and manipulation using a combination of various losses, we consider a simpler approach leading to surprisingly strong results.
We show that texture loss\cite{gatys2016image} alone allows the generation of perceptually high quality images. We provide a better understanding of texture constraining mechanism and develop a novel semantically guided texture constraining method for further improvement. Using a recently developed perceptual metric employing  ``deep features'' and termed LPIPS \cite{zhang2018unreasonable}, the method obtains state-of-the-art results. Moreover, we show that a texture representation of those deep features better capture the perceptual quality of an image than the original deep features. Using  texture information, off-the-shelf deep classification networks (without training) perform as well as the best performing (tuned and calibrated) LPIPS metrics. The \href{https://github.com/waleedgondal/Texture-based-Super-Resolution-Network}{code} is publicly available.
\end{abstract}

\section{Introduction}
\label{sec:introduction}

Recently, the task of single image super-resolution (SISR) has taken an interesting turn. Convolutional neural networks (CNNs) based models have not only been shown to reduce the distortions on full reference (FR) metrics for, e.g., PSNR, SSIM and IFC \cite{johnson2016perceptual,ledig2016photo,lim2017enhanced,LapSRN,dong2014learning,kim2010single}, but also to produce perceptually better images \cite{ledig2016photo,SajSchHir17}. The models trained specifically to reduce distortions fail at producing visually compelling results. They suffer from the issue of ``regression-to-the-mean'' as they mainly rely on minimizing the mean square error (MSE) between a high resolution image $I_{HR}$ and an estimated image $I_{est}$, approximated from its low resolution counterpart $I_{LR}$. This minimization of MSE leads to the suppression of high frequency details in $I_{est}$, entailing blurred and over-smoothed images. Therefore, FR metrics do not conform with the human perception of visual quality. This was illustrated in \cite{laparra2016perceptual,wang2004image} and recently mathematically analyzed in \cite{blau2017perception}.

The recently proposed methods \cite{ledig2016photo,SajSchHir17,wang2018recovering} made substantial progress in improving the perceptual quality of the images by building on generative adversarial networks (GANs) \cite{goodfellow2014generative}. The adversarial setting of a generator and a discriminator network helps the generator in hallucinating high frequency textures into the resultant images. Since the goal of the generator is to fool the discriminator, it may hallucinate fake textures which are not entirely faithful to the input image. This fake texture generation can be clearly observed in an $8\times$ image super-resolution task. This behavior of GANs can be reduced using a combination of content preserving losses. This not only limits the ability of the generator to induce high quality textures but also makes it fall short in reproducing image details in the regions which have complex and irregular patterns such as tree leaves, rocks etc.

\begin{figure}[t]
	\centering
	\begin{tabular}{c@{\hspace{0.01\linewidth}}c@{\hspace{0.01\linewidth}}c@{\hspace{0.01\linewidth}}c@{\hspace{0.01\linewidth}}c@{\hspace{0.01\linewidth}}c@{\hspace{0.01\linewidth}}c}
		\includegraphics[width = .154\linewidth]{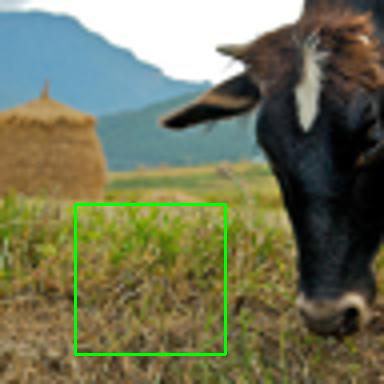} &		
		\includegraphics[width = .154\linewidth]{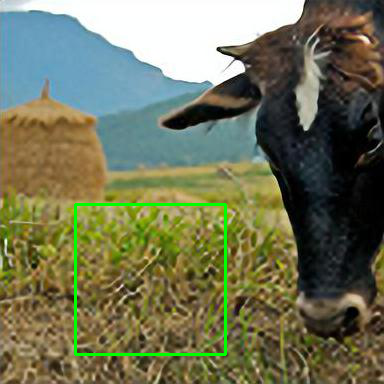} &
		\includegraphics[width = .154\linewidth]{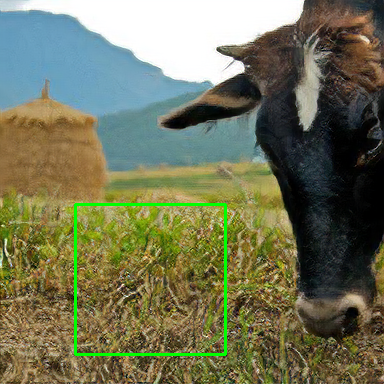} &
		\includegraphics[width = .154\linewidth]{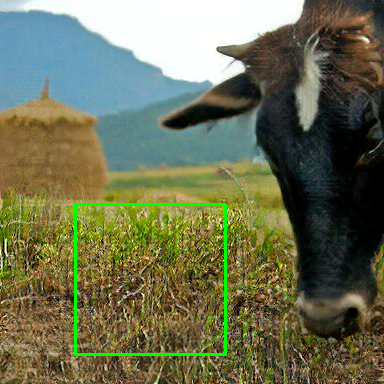} &
		\includegraphics[width = .154\linewidth]{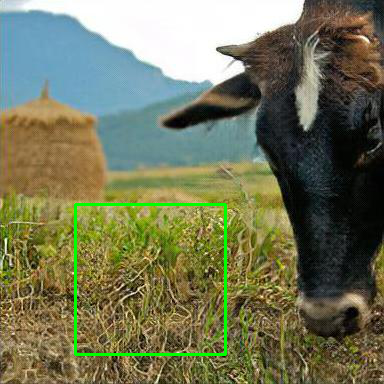} &
		\includegraphics[width = .154\linewidth]{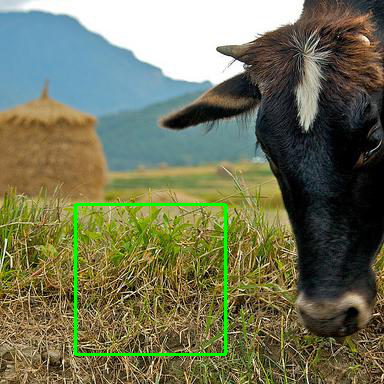} &\\
		\includegraphics[width = .154\linewidth]{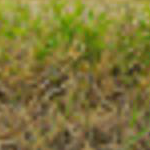} &
		\includegraphics[width = .154\linewidth]{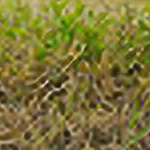} &
		\includegraphics[width = .154\linewidth]{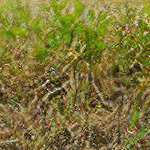} &
		\includegraphics[width = .154\linewidth]{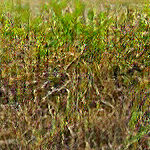} &
		\includegraphics[width = .154\linewidth]{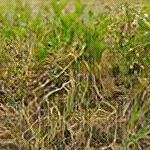} &
		\includegraphics[width = .154\linewidth]{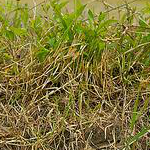} &\\

		{(a) Bicubic }& {(b)SRresnet\cite{ledig2016photo}} & {(b)ENet \cite{SajSchHir17}} & {(c) SRGAN\cite{ledig2016photo}} & {(d)TSRN(ours)} & {(e) Original } \\
	\end{tabular}
	
	\caption{Visual Comparison of the recent state-of-the-art methods as measured by distortion and perceptual quality metrics with our texture based super-resolution network (TSRN) for 4$\times$ SISR.}
	\label{fig:third}
\end{figure}
In the present paper we show that, in the task of SISR, perceptually high quality textures can be synthesized on the estimated images $I_{est}$ using the Gram matrices based texture loss \cite{gatys2016image}. The loss was first employed by Gatys et al. in transferring realistic textures from a style image ($I_s$) to a content image ($I_c$). Despite the success of this method, the utility of texture transfer for enhancing natural images has not been studied extensively. This is because of the fact that while preserving the local spatial information of the textures, the texture loss discards the global spatial arrangement of the content image, rendering the semantic guidance of texture transfer a difficult problem.

We explore the effectiveness of Gram matrices in transferring and hallucinating realistic texture in the task of SISR. We show that despite its simplicity through the use of a single loss function, our proposed network yields favorable results when compared to state-of-the-art models that employ a mixture of loss functions and involve GANs that are notoriously difficult to train. In contrast, our model converges without the need of hand-tuned training schemes. We further build on this finding by providing external semantic guidance to control the texture transfer. We show that this scheme prevents the random spread of small features across object boundaries thus improving the visual quality of results especially in the challenging task of 8$\times$ SISR. Furthermore, we demonstrate, that Gram matrices of deep features perform surprisingly well in measuring human perceived similarity between image patches.

\section{Related Work}
\label{sec:relatedwork}

\subsection{Super Resolution}
Single image super-resolution(SISR) is the problem of approximating
a high resolution ($I_{HR}$) image from its corresponding low resolution ($I_{LR}$) input image. The task is to fill in missing information in $I_{HR}$ which involves the reconstruction and hallucination of textures, edges and low-level image statistics while remaining faithful to the low-resolution $I_{LR}$ input. It is an under-determined inverse problem where different image priors have been explored to guide the upsampling of $I_{LR}$ \cite{zoran2011learning,sun2008image,liu2016robust}. However, much success has been achieved by using recent data-driven approaches where a large number of training examples are used to set the prior over the empirical distribution of data. One of the earliest methods involved simple interpolation schemes \cite{hou1978cubic}, e.g. bicubic, Lanczos. Due to their simplicity and fast inference, these methods have been widely used, however they suffer from blurriness and can not predict high frequency details.

In recent years, learning based methods that try to learn a mapping between $I_{LR}$ to $I_{HR}$ have enjoyed much attention and progress. These methods can be classified into parametric and non-parametric methods~\cite{huang2015single}. Non-parametric algorithms include neighborhood embedding algorithms~\cite{freeman2002example,chang2004super,timofte2014a+,yang2013fast}, that seek for the nearest match in an available database and try to synthesize an image by simple blending of different patches. Prone to mismatch and misalignment in patches these methods suffer from rendering artifacts in the HR output \cite{sun2012super}. Parametric methods include sparse models \cite{liu2016robust}, regression functions \cite{kim2010single} and convolutional neural networks (CNNs). Dong et al. \cite{dong2014learning} first employed a shallow CNN to perform SISR on a bicubic interpolated image and got impressive results, \cite{kim2016accurate} successfully used a deep residual network.
These CNN based methods use mean square error (MSE) as an optimization objective which leads to blurriness and fails to reconstruct high frequency details.  Methods like \cite{johnson2016perceptual,ledig2016photo} tend to overcome this issue by minimizing perceptual losses in feature space. Ledig et al. \cite{ledig2016photo} proposed SRResNet to show improvements in full-reference (FR) metrics. Follow-up work used a multi-scale optimized SRResNet architecture to win the NTIRE 2017 Super-Resolution Challenge \cite{timofte2017ntire} for 4x super-resolution. Moreover, \cite{LapSRN} uses a coarse-to-fine laplacian pyramid framework to achieve state-of-the-art results in 8x super-resolution with respect to FR metrics.

More recently, GANs based methods \cite{ledig2016photo,SajSchHir17,wang2018recovering} showed promising results by drastically improving the perceptual quality of images. In addition to the perceptual and adversarial losses used by \cite{ledig2016photo}, the patch-wise texture loss used by \cite{SajSchHir17} helps synthesizing high quality textures. Our approach is different from \cite{SajSchHir17}, as we give up on the adversarial and perceptual loss terms. Moreover, we also don't use patch-wise texture loss and show that a globally applied texture loss is enough for spatially aligning textures and generating photo-realistic high-quality images. \cite{sun2017super} also used patches and manually derived segmentation masks to constrain the texture synthesis in $I_{est}$. However, it highly relies on the efficiency of a slow patch matching algorithm and thus is prone to wrong matching of regions in $I_{est}$ and $I_{HR}$ which renders artifacts. The loss is also shown to be an important ingredient of a recent image-inpainting method \cite{liu2018image}. A new deep features based contextual loss \cite{mechrez2018contextual} is used by \cite{mechrez2018Learning} to maintain the natural image statistics of $I_{est}$.

\subsection{Neural Texture Transfer}
The concept of neural texture transfer was first coined by Gatys et al. \cite{gatys2016image}. The method relies on matching the Gram matrices of VGG-19 \cite{simonyan2014very} features to transfer the texture of one image to another. 
Afterwards, much work has been done in order to improve the speed \cite{johnson2016perceptual,ulyanov2016texture} and quality \cite{luan2017deep,yeh2018improved,liao2017visual} of style transfer using feed forward networks and perceptual losses. Building on fast style transfer, \cite{dumoulin2016learned,chen2017stylebank} proposed models to transfer textures from multiple style images. \cite{yeh2018improved} showed improvement in style transfer by computing cross-layer Gram matrices instead of within-layer Gram matrices. Recently, Li et al. \cite{li2017demystifying} has shown that matching the Gram matrices for style transfer is equivalent to minimizing MMD with the second order polynomial kernel. In addition to improving the style transfer mechanism, some work has been done to spatially constrain the texture transfer in order to maintain the textural integrity of different regions\cite{lu2017decoder,gatys2016controlling}. Gatys et al. \cite{gatys2016controlling} demonstrated the spatial control of texture transfer using guided Gram matrices where binary masks are used as guidance channels in order to constrain the textures. Similar scheme was used by \cite{luan2017deep} in constraining style transfer. Instead of enforcing spatial guidance in the feature space of deep networks like these methods, we enforce it in pixel-space via customized texture loss which, unlike other methods, not only enables it to easily scale to multiple style images but also does not require semantic details at the test time.

\noindent Our main contributions are as follows:
\begin{itemize}[leftmargin=*]
\itemsep0em 

\item We provide a better understanding of texture constraining mechanism via texture loss and show that SISR of high perceptual quality can be achieved by using this as an objective function. The results compare well with GANs based methods on 4x SISR and outperform them on 8x SISR. 
\item Unlike GANs based methods, our method is easily reproducible and generates faithful textures especially in the constrained domain of facial images. 

\item To further enhance the quality of 8x SISR results, we formulate a novel semantically guided texture transfer scheme in order to avoid the intermixing of interclass textures such as grass, sky etc. The method is easily scalable to multiple style images and does not require semantic details at test time. 

\item We also show that Gram matrices provide a better and richer framework to capture the perceptual quality of images. Using this, our off-the-shelf deep classification networks (without training) perform as well as the best performing (tuned and calibrated) LPIPS metrics \cite{zhang2018unreasonable}.

\end{itemize}

\section{Texture Loss}
The texture transfer loss was first proposed in the context of neural style transfer~\cite{gatys2016image}, where both style $I_{s}$ and content images $I_{c}$ are mapped into feature space using a VGG-19 architecture \cite{simonyan2014very}, pre-trained for image classification on image-net. The feature maps of both $I_{s}$ and $I_{c}$ are denoted by $F^l \in \mathbb{R}^{N_l \times M_l}$ and $P^l \in \mathbb{R}^{N_l \times M_l}$ respectively, where $N_l$ is the number of feature maps in layer \textit{l} and $M_l$ is the product of height and width of feature maps in layer \textit{l} i.e. $M_l =  height \times width$. A Gram matrix is the inner product of vectorized feature maps. Therefore the Gram matrices for both $F^l$ and $P^l$ are computed as 
$G_{i,j}^l = \textbf{F}^{T}_{i} \textbf{F}_{j} $ and $A_{i,j}^l = \textbf{P}^{T}_{i} \textbf{P}_{j}$. The texture loss $\mathcal{L}_{texture}$ is defined by the mean squared error between the feature correlations expressed by these Gram matrices.
\begin{equation}
\mathcal{L}_{texture} = \frac{1}{4N_l^2M_l^2} \sum_{i=1}^{N_l} \sum_{j=1}^{M_l} (G_{i,j}^l - A_{i,j}^l)^2
\label{eqtn:loss}
\end{equation}
The loss tries to match the global statistics of $I_{c}$ with $I_{s}$, captured by the correlations between feature responses in layers \textit{l} of the VGG-19. These correlations capture the local spatial information in the feature maps while discard their global spatial arrangement \cite{gatys2015texture}.

\subsection{Constraining Texture Transfer}
\begin{figure}[h!]
	\includegraphics[width=\linewidth]{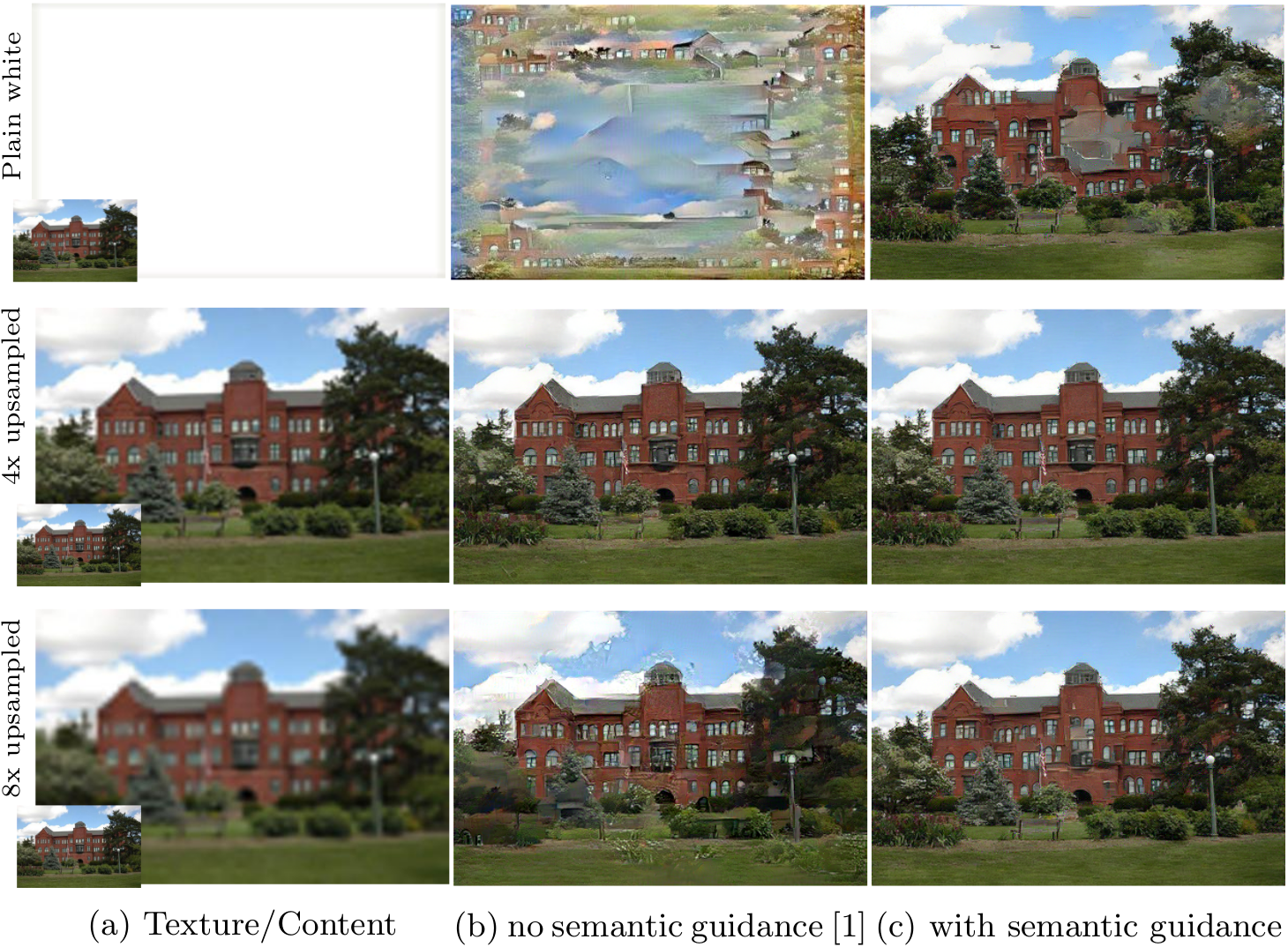}
	\caption{(a) shows $I_{HR}$ (in insets) and a plain white, 4x and 8x upsampled versions of $I_{HR}$ as $I_{c}$. (b) vanilla neural texture transfer \cite{gatys2016image}. (c) neural texture transfer with semantic guidance.}
	\label{fig:texture_transfer}
\end{figure}

The above loss tries to match the global level statistics of $I_{s}$ and $I_{c}$ without retaining the spatial arrangement of the content image. However, we observe that if there exists a good feature space correspondence between $I_{s}$ and $I_{c}$ then the Gram matrices alone constrain the texture transfer such that it preserves the semantic details of the content image. The composition of Gram matrices makes use of the translational invariance property of the pre-trained VGG-19's \cite{simonyan2014very} convolutional kernels in mapping the textures correctly. We shed more light on this texture constraining mechanism and its translational invariant mapping in the appendix. Thus Gram matrices' provide a stable spatial control such that the texture from $I_{s}$ maps to the corresponding features on $I_{c}$. Fig \ref{fig:texture_transfer} shows texture transfer of a non-texture image for different initial approximates of $I_{c}$ using iterative optimization approach by \cite{gatys2016image}. Second column depicts the results of vanilla style transfer \cite{gatys2016image} on a plain white image, 4x upsampled image and an 8x upsampled images respectively. In case of plain white image, the texture gets transferred in an uncontrollable fashion. This is the known phenomenon in image style transfer. However, the texture transfer on a 4x and 8x upsampled images shows consistency in texture mapping i.e. texture from $I_{s}$ gets mapped to the correct corresponding regions of $I_{c}$. We observe that the interpolated approximates $I_{est}$ of $I_{LR}$ are good enough for establishing feature-space correspondences and thus mapping the textures correctly.

 In the Fig \ref{fig:texture_transfer}, one can observe that the texture transfer for a 4x interpolated image is much better than that for an 8x. The ambiguousness in texture transfer for an 8x upsampled $I_{LR}$ is because of the absence of enough content features to establish correspondences. Thus to better guide the texture transfer in 8x SISR, we devise an external semantic guidance scheme. The third column in Fig \ref{fig:texture_transfer} shows the effectiveness of the semantically guided texture transfer. In comparison to the second column we can see that the texture is transferred in a more coherent fashion.
\subsection{Texture Loss in SISR.}
\label{constrain_texture}
 
In SISR we try to find a mapping between a low-resolution input image $I_{LR}$ and a high-resolution output image $I_{HR}$. As a function approximator we use a deep CNN. While recent state-of-the-art methods use a combination of various loss functions, our texture super resolution network (TSRN) is specifically trained to optimize for $\mathcal{L}_{texture}$ in equation \ref{eqtn:loss} which yields images of perceptually high quality for $4\times$ and $8\times$ super-resolution, Fig \ref{fig:first} and \ref{fig:fourth}.

\subsection{SISR via Semantically Constrained Textures.}
In order to make full use of the texture loss based image super resolution, we also performed externally controlled semantic texture transfer. We enforce semantic details via loss function. For the implementation of semantic control of texture transfer, we use the ground truth segmentation masks provided by the recently released dataset MS-COCO stuff dataset \cite{lin2014microsoft}.

Additional spatial control is provided by making use of the semantic information present inside an image. Instead of matching the global level statistics of an image we divide the image into \textit{r} segments semantically. Each segment exhibits its own local level statistics which are different from the other segments of the same image. This facilitates us to match the local level statistics at an individual segment level. Also it helps in preserving the global spatial arrangement of the segments as the relative spatial information of each segment is considered before extracting them from the images.

\noindent Our method gains inspiration from the spatial control of texture transfer based on guided Gram matrices (GGMs) \cite{gatys2016controlling} where binary segmentation masks are used to define which region of a style image would get mapped to the specific region of a content image. It uses \textit{r} segmentation masks $I_{seg}^r$ to compute guidance channels $(\textbf{T}_l^r)$ for each layer \textit{l} of a CNN by either down-sampling them to match the dimensions of each layer's feature maps or by enforcing spatial guidance only on neurons whose receptive field lie inside the guidance region for better results. The guidance channels are then used to form spatially guided feature maps by the element-wise multiplication of texture image features and the guidance channels. This method of computing GGMs for training a deep architecture is not feasible, especially in our case where we have multiple segmentation masks for each image. As the computation of spatially guided feature maps depends on the element-wise multiplication of $(\textbf{T}_l^r)$ with texture image feature maps which depending on the size and depth of chosen CNN layers can be computationally very expensive. Moreover, the computation of better guidance channels require extra efforts of discarding the neurons whose receptive fields lie outside of guidance region.\\
We propose a simplification of this process by removing the need of guidance channels $(\textbf{T}_l^r)$ and the explicit computation of spatially guided feature maps altogether. The \textit{r} binary segmentation masks $I_{seg}^r$ (having pixel value of 1 for the class of interest and 0 elsewhere) where each mask categorically

\begin{figure}[h!]
	\includegraphics[width=\linewidth]{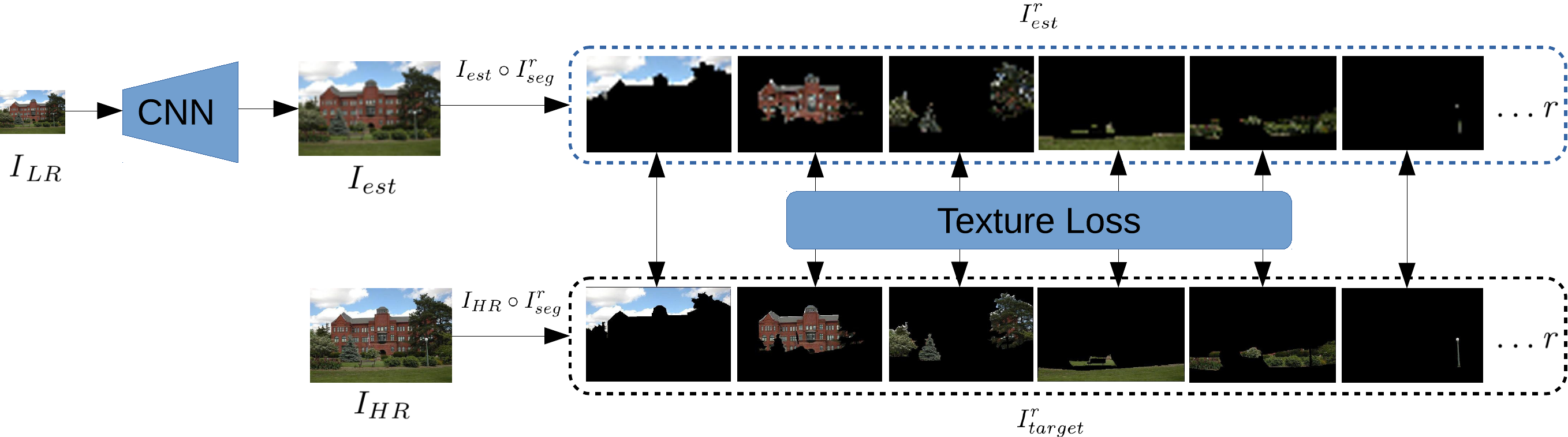}
	\caption{Scheme for semantically controlled texture transfer.}
	\label{fig:method}
\end{figure}

 represents a different region of an image are element-wise multiplied with the texture image $I_{HR}$ and the estimated image $I_{est}$ to give out $I_{target}^r$ and $I_{est}^r$ respectively, Fig \ref{fig:method}.

\begin{equation}
I_{target}^r = I_{HR} \circ I_{seg}^r
\label{target_masks}
\end{equation}
\begin{equation}
I_{est}^r = I_{est} \circ I_{seg}^r
\label{est_masks}
\end{equation} 

These segmented images are then propagated to the VGG19 and Gram matrices of their feature maps are then computed in normal fashion. The method is flexible and relatively fast to enforce spatial guidance of texture transfer, especially when it has to be used for training a deep architecture. The texture loss is then performed individually for all the segmented images. Equation \ref{loss_semantica} shows the objective function formulation of the complete semantically controlled texture transfer. See abstract to check the effectiveness of our proposed semantically controlled fast style transfer.

\begin{equation}
\mathcal{L}_{texture} = \sum_{k=1}^{r} \frac{1}{4N_l^2M_l^2} \sum_{i=1}^{N_l} \sum_{j=1}^{M_l} (G_{i,j}^l(I_{target}^k) - A_{i,j}^l(I_{est}^k))^2
\label{loss_semantica}
\end{equation}

\section{Architecture}
For the implementation of TSRN, we employ a fully convolutional neural network architecture inspired by \cite{SajSchHir17}. The architecture is efficient at inference time as it performs most feed forward computations on $I_{LR}$ and is deep enough to perform texture synthesis. The presence of residual blocks facilitates convergence during training. Similarly to \cite{SajSchHir17}, we also add a bi-cubically upsampled verison of $I_{LR}$ to the predicted output such that the network is only required to learn the residual image. This helps to reduce color shifts during training as also reported by \cite{SajSchHir17}. However, instead of using nearest neighbor up-sampling, we use a pixel resampling layer \cite{shi2016real} because of its recent proven success in generative networks \cite{karras2017progressive}. The method is also shown to be agnostic to model's depth. See appendix for more details.

\section{Implementation}
We trained our network on MS-COCO \cite{lin2014microsoft}, where we center crop image patches sized 256$\times$256 pixels. The patches are then bi-cubically down-sampled 4$\times$ or 8$\times$ to 64$\times$64 or 32$\times$32, respectively. We first pretrain our network by minimizing mean square error (MSE) for 10 epochs. We found this pre-training beneficial for the subsequent Gram matrix based optimization as it facilitates the detection of relevant features for texture transfer. After pretraining, we train our model using only \ref{eqtn:loss}
\begin{figure}[h!]
	\includegraphics[width=\linewidth]{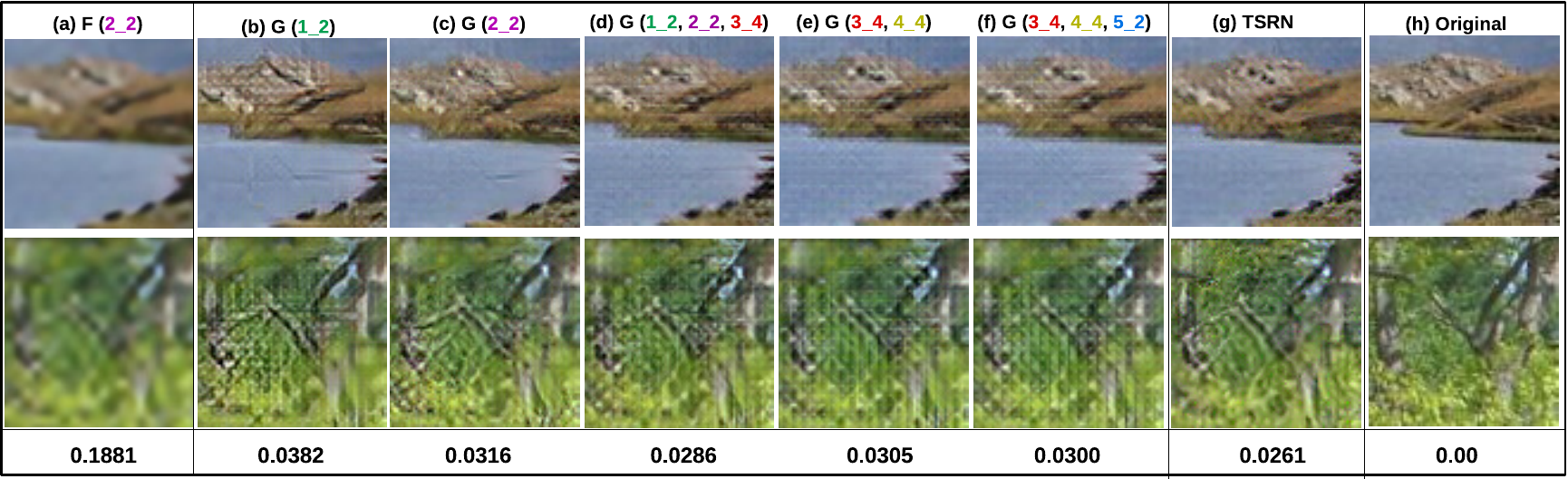}
	\caption{Layer and loss ablation study on SunHays dataset \cite{sun2012super}. Each column shows the effects of different VGG19\cite{simonyan2014very} layers on the visual quality of a restored image. Perceptual loss using deep features (F) generates blurred images (left most column) in comparison to Gram matrices (G) based restoration. The last row shows the mean LPIPS score on the dataset (lower score is better).}
	\label{ablation}
\end{figure}

  as an objective function for another 100 epochs. We found that the network converges after approximately 60 epochs. For the implementation of $\mathcal{L}_{texture}$, we compute Gram matrices on layers \textit{conv2\_2}, \textit{conv3\_4}, \textit{conv4\_4} and \textit{conv5\_2} of a pre-trained VGG-19 architecture. To justify the selection of specific VGG-19's layers for texture loss, we provide a qualitative and quantitative (LPIPS) analysis on SunHays dataset in \figurename{\ref{ablation}}. We considered convolutional layers before each pooling layer except (\textit{conv1\_2}) as this layer, containing more pixel-level and less structural information, causes artifacts and over-smoothing in images. The selection of only higher layers tend to generate checkboard artifacts. In \figurename{\ref{ablation}}, all the networks are trained using the same architecture and procedure mentioned in the paper for 100 epochs. The network is trained with the learning rate of 0.0005 using ADAM as an optimizer. We use the PyTorch framework \cite{paszke2017automatic} to implement the model on a Nvidia Tesla P40 GPU. Inference time for 4$\times$ and 8$\times$ SISR is approximately 41 and 32 milliseconds for a 1 mega-pixel image and 0.203 and 0.158 seconds for a 5 mega-pixel image on the GPU.\\

For our results on segmentation based super-resolution (TSRN-S), we pre-train on the MS-COCO dataset before we train on the MS-COCO stuff dataset using equation \ref{loss_semantica} as an objective function. The stuff dataset is particularly suited for our task as it not only contains the segmentation masks of object instances but also outdoor scenes like grass, sky, buildings etc. Statistically, these regions cover more than 60\% \cite{zhou2016semantic}
of images showing natural scenes. To reduce the computation time, we consider the binary segmentation masks of only six maximally represented classes in each image (based on their pixel count). Whereas the seventh mask covers the 'others' class, containing the remaining regions of the image. If there are less than six classes in an image then the 'others' class is replicated to give out seven masks per image. Both code and pre-trained models are publicly available.

\section{Experimental Results}
We evaluate both our proposed models, one with globally computed Gram matrices (TSRN-G) and semantically guided Gram matrices (TSRN-S). We provide extensive quantitative and qualitative comparisons with recent state-of-the-art methods in SISR as measured by distortion metrics (PSNR, SSIM etc.) but also in terms of perceptual quality. 

\subsection{Quantitative Evaluation}
Distortion metrics have been shown to not correlate
well with perceived image
quality~\cite{SajSchHir17,zhang2018unreasonable}. Therefore, for
quantitative comparison we follow \cite{SajSchHir17} and report the
performance in object recognition as a proxy for perceived image
quality. Additionally, we report numbers for a recently proposed
learned full-reference image quality metric
\cite{zhang2018unreasonable} that approximates perceptual similarity.

\subsubsection{Object Recognition Performance.}
The perceptual quality of an image correlates very well with its performance on object recognition models which are trained on the large corpus of image-net, as corroborated by \cite{SajSchHir17}. Recently, the same methodology of assessing image quality has been adopted by a competition\footnote{http://www.ug2challenge.org/}. The idea stems from the observation that deep features are very effective in capturing the perceptual quality of images~\cite{zhang2018unreasonable}. Therefore, we perform our comparison with other methods utilizing the standard image classification models trained on ImageNet. We randomly pick 1000 images from the ILSVRC 12 validation
\begin{table*}[h]
	\centering
	\resizebox{\linewidth}{!} {
		\begin{tabular}{l l@{\hskip .05in} c@{\hskip .05in} c@{\hskip .05in} c@{\hskip .15in} c@{\hskip .05in} @{\hskip .05in}c@{\hskip .05in} c@{\hskip .05in} c@{\hskip .05in} c@{\hskip .15in} c@{\hskip .05in} } \toprule
			
			\textbf{TopK} & \textbf{Methods} & \textbf{Bicubic} & \textbf{SRResNet\cite{ledig2016photo}} & \textbf{SRGAN\cite{ledig2016photo}} & \textbf{ENet-PAT\cite{SajSchHir17}} & \textbf{TSRN-S} & \textbf{TSRN-G} & \textbf{Baseline} \\ \midrule
			
			\multirow{3}{*}{Top 1}& DenseNet-169 & 0.594 & 0.641 &0.666 & 0.658 &0.688 &\textbf{0.692} & 0.713\\
			& ResNet-50  & 0.545 & 0.616 &0.655 & 0.649 & \textbf{0.674}&0.671 & 0.703  \\
			& VGG-19  & 0.455 & 0.538 &0.578 & 0.571 &\textbf{0.610}& 0.609 & 0.656 \\ \midrule
			\multirow{3}{*}{Top 5}& DenseNet-169  & 0.788 & 0.862 &0.864 & 0.857 & \textbf{0.876}&0.871 & 0.890\\
			& ResNet-50  & 0.776 & 0.841 &0.847 & 0.843 &0.862& \textbf{0.866} & 0.885 \\
			& VGG-19   & 0.676 & 0.772 &0.798 & 0.792 &0.819& \textbf{0.821}& 0.853 \\				
			\bottomrule
		\end{tabular}
	}
	\caption{ Top-1 and Top-5 image recognition accuracy on $4\times$ SISR images.}
	\label{tab:top_4x}
	\vspace{-2mm}
\end{table*}

 dataset and super-resolve their downsampled versions using different super-resolution models. The performance is evaluated on how much recognition accuracy is retained by each model, compared to the baseline accuracy. Tables~\ref{tab:top_4x} and \ref{tab:top_8x} show that our proposed TSRN model outperforms all other state-of-the-art SISR methods for both 4$\times$ and 8$\times$ super-resolution.

\begin{table*}[t]
	\centering
	\resizebox{\linewidth}{!} {
		\begin{tabular}{l l@{\hskip .05in} c@{\hskip .05in} c@{\hskip .05in} c@{\hskip .15in} c@{\hskip .05in} @{\hskip .05in}c@{\hskip .05in} c@{\hskip .05in} c@{\hskip .05in} c@{\hskip .15in} c@{\hskip .05in} } \toprule
			\textbf{TopK} & \textbf{Methods} & \textbf{Bicubic} & \textbf{SRResNet\cite{ledig2016photo}} & \textbf{SRGAN\cite{ledig2016photo}} & \textbf{TSRN-S} & \textbf{TSRN-G} & \textbf{Baseline} \\ \midrule
			
			\multirow{3}{*}{Top 1}&DenseNet-169 & 0.353&0.506  & 0.432 &\textbf{0.509}& 0.506 & 0.713 \\
			&ResNet-50  & 0.301 &0.437 & 0.424 &0.484& \textbf{0.503} & 0.703   \\
			&VGG-19  & 0.239  &0.343& 0.267 &0.374& \textbf{0.389} & 0.656 \\ \midrule
			\multirow{3}{*}{Top 5}&DenseNet-169  & 0.602 &0.727& 0.676&0.733 & \textbf{0.743} & 0.890\\
			&ResNet-50  & 0.518 &0.689 &0.657 &\textbf{0.718}& 0.717 & 0.885 \\
			&VGG-19   & 0.406 &0.565 &0.504 & \textbf{0.613}&0.611& 0.853 \\			
			\bottomrule
		\end{tabular}
	}
	\caption{ Top-1 and Top-5 image recognition accuracy on $8\times$ SISR images.}
	\label{tab:top_8x}
	\vspace{-2mm}
\end{table*}

\subsubsection{LPIPS.}
The Learned Perceptual Image Patch Similarity (LPIPS) metric \cite{zhang2018unreasonable} is a recently introduced full-reference image quality assessment metric which tries to measure the perceptual similarity between two images. The metric uses linearly calibrated off-the-shelf standard deep classification networks trained to measure the perceptual similarity of the images. The networks are trained on the very large Berkeley-Adobe Perceptual Patch Similarity (BAPPS) \cite{zhang2018unreasonable} dataset, containing human perceptual judgments. We use the pre-trained, linearly calibrated AlexNet and SqueezeNet networks\footnote{https://github.com/richzhang/PerceptualSimilarity}. The networks are trained on patches sized 64$\times$64 pixels. Therefore, we also divide the images into patches of size 64$\times$64 pixels. For each image, we pick its shorter dimension and find the nearest possible value \textit{v} divisible by 64, then we center crop an image of resolution \textit{v} $\times$ \textit{v}. The cropped image is then further divided into patches of size 64 $\times$ 64. We report the averaged perceptual similarity determined on those patches.\\
In \tablename{~\ref{comp_1}} we use the recommended AlexNet (linear) and SqueezeNet (linear) models for measuring the perceptual quality. We found the quantitative evaluations to be consistent across numerous models that have been trained to improve either PSNR, SSIM scores such as SRResNet, LapSRN, SRCNN or the ones trained to improve perceptual quality such as SRGAN and ENet-PAT.  TSRN consistently achieves better perceptual similarity scores than other methods.

\begin{table*}[t]
	\begin{center}
		\resizebox{\linewidth}{!} {
			\begin{tabular}{l@{\hskip .05in} c@{\hskip .05in} c@{\hskip .05in} c@{\hskip .15in} c@{\hskip .05in} @{\hskip .05in}c@{\hskip .05in} c@{\hskip .05in} c@{\hskip .05in} c@{\hskip .15in}} \toprule
				\centering
				\multirow{2}[3]{*}{\textbf{Metric}} & \multicolumn{2}{c}{\textbf{Set 5}} & \multicolumn{2}{c}{\textbf{Set 14}} &
				\multicolumn{2}{c}{\textbf{BSD 100}} &
				\multicolumn{2}{c}{\textbf{Urban}}\\ 
				\cmidrule(lr){2-3} \cmidrule(lr){4-5} \cmidrule(lr){6-7} \cmidrule(lr){8-9}
				\centering
				& \textbf{AlexNet} & \textbf{SNet} & \textbf{AlexNet} & \textbf{SNet}  & \textbf{AlexNet} & \textbf{SNet} & \textbf{AlexNet} & \textbf{SNet} \\ \midrule
				
				Bicubic & 0.1585 & 0.1202 & 0.1731& 0.1320 & 0.1463 &0.1007 & 0.1552& 0.1238 \\
				SRCNN\cite{dong2014learning} & 0.0964 & 0.0732 & 0.1175 & 0.1025 & 0.1257 & 0.0920 & 0.0960 & 0.0905\\
				LapSRN\cite{LapSRN} & 0.0566 & 0.0556 & 0.1002& 0.0967 & 0.1005 &0.0753 & 0.0746& 0.0757 \\
				MSLapSRN\cite{MSLapSRN} & 0.0551 & 0.0574 & 0.0972& 0.0916 & 0.0989 &0.0720 & 0.0691& 0.0709 \\				
				SRResNet\cite{ledig2016photo} & 0.0538 & 0.0491 & 0.0848& 0.0821 & 0.0909 &0.0625 &  0.0628&0.0652 \\
				SRGAN\cite{ledig2016photo}  & 0.0275& 0.0466 & 0.0575& 0.0679 &  0.0484 &0.0527 & 0.0401& 0.0584\\
				ENet-PAT\cite{SajSchHir17} & \textbf{0.0251} & 0.0391 & 0.0569&0.0590 & 0.0494&0.0472 & 0.0414&0.0467\\	
				\midrule
				TSRN-S (Ours) &0.0273&0.0394&\textbf{0.0438}& 0.0483& \textbf{0.0478}&0.0420&0.0397&0.0404\\  
				TSRN-G (Ours) & 0.0285&\textbf{0.0358} & 0.0463& \textbf{0.0456} & 0.0481 &\textbf{0.0404} & \textbf{0.0385}& \textbf{0.0392}\\  
				\bottomrule
			\end{tabular}
		}
		\vspace{1mm}
		\caption{Comparison for 4$\times$ SISR on pre-trained AlexNet-linear and SqueezeNet-linear LPIPS metric\cite{zhang2018unreasonable}. Lower score is better.}
		\label{comp_1}
		\vspace{-4mm}
	\end{center}
\end{table*}

\begin{table*}[t]
	\centering
	\resizebox{\linewidth}{!} {
		\begin{tabular}{l@{\hskip .05in} c@{\hskip .05in} c@{\hskip .05in} c@{\hskip .15in} c@{\hskip .05in} @{\hskip .05in}c@{\hskip .05in} c@{\hskip .05in} c@{\hskip .05in} c@{\hskip .15in}} \toprule
			\centering
			\multirow{2}[3]{*}{\textbf{Metric}} & \multicolumn{2}{c}{\textbf{Set 5}} & \multicolumn{2}{c}{\textbf{Set 14}} &
			\multicolumn{2}{c}{\textbf{BSD 100}} &
			\multicolumn{2}{c}{\textbf{Urban}}\\ 
			\cmidrule(lr){2-3} \cmidrule(lr){4-5} \cmidrule(lr){6-7} \cmidrule(lr){8-9}
			\centering
			& \textbf{AlexNet} & \textbf{SNet} & \textbf{AlexNet} & \textbf{SNet}  & \textbf{AlexNet} & \textbf{SNet} & \textbf{AlexNet} & \textbf{SNet} \\ \midrule
			
			Bicubic & 0.27464 & 0.22877 & 0.27390 & 0.24669 & 0.22802 & 0.20202 & 0.23854 & 0.22946 \\
			LapSRN\cite{LapSRN} & 0.19849 & 0.15506 & 0.21525 & 0.19058 & 0.19009 & 0.16379 & 0.15638 & 0.15426\\
			MSLapSRN\cite{MSLapSRN} & 0.16748 & 0.13609 & 0.20184 & 0.17599 & 0.17679 & 0.15276 & 0.13252 & 0.13328 \\
			SRResNet\cite{ledig2016photo} & 0.13679 & 0.11958 & 0.18091 & 0.16060 & 0.16148 & 0.13512 &  0.13714 & 0.13217 \\
			SRGAN\cite{ledig2016photo}  & 0.14230 & 0.15007 & 0.13801 & 0.12720 &  0.13276 & 0.10902 & 0.12929 & 0.12470\\
			\midrule
			TSRN-S(Ours) & \textbf{0.0859} & 0.0863 & \textbf{0.1194} & \textbf{0.0963}  &\textbf{0.1021} & \textbf{0.0823} &0.0918 &\textbf{0.0802} \\ 
			TSRN-G(Ours) & 0.0900 & \textbf{0.0859} & 0.1277& 0.1092 & 0.1029 & 0.0833 & \textbf{0.0900}& 0.0817\\  
			
			\bottomrule
		\end{tabular}
	}
	\vspace{1mm}
	\caption{Comparison for 8$\times$ SISR on pre-trained AlexNet-linear and SqueezeNet-linear LPIPS Perceptual Similarity Metric models. Lower score is better.}
	\label{comp_2}
	\vspace{-3mm}
\end{table*}
\subsection{Visual Comparison}
\begin{figure}[t]
	\centering
	
	\begin{tabular}{c@{\hspace{0.01\linewidth}}c@{\hspace{0.01\linewidth}}c@{\hspace{0.01\linewidth}}c@{\hspace{0.01\linewidth}}c@{\hspace{0.01\linewidth}}c@{\hspace{0.01\linewidth}}c}
		
		\includegraphics[width = .153\linewidth]{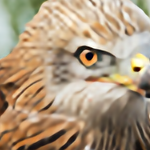} &
		\includegraphics[width = .153\linewidth]{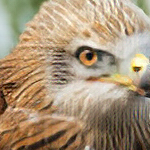} &
		\includegraphics[width = .153\linewidth]{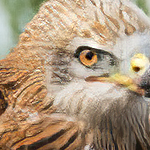} &
		\includegraphics[width = .153\linewidth]{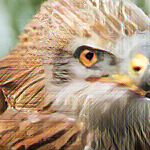} &
		\includegraphics[width = .153\linewidth]{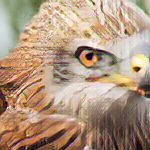} &
		\includegraphics[width = .153\linewidth]{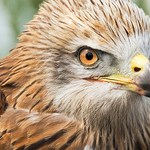} \\
		
		\includegraphics[width = .153\linewidth]{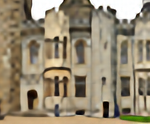} &
		\includegraphics[width = .153\linewidth]{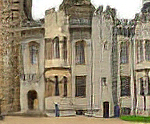} &
		\includegraphics[width = .153\linewidth]{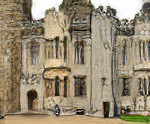} &
		\includegraphics[width = .153\linewidth]{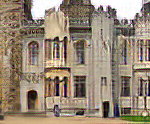} &
		\includegraphics[width = .153\linewidth]{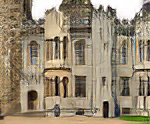} &
		\includegraphics[width = .153\linewidth]{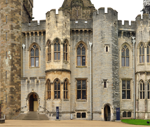} \\

		{(a) SRResNet } & {(b)SRGAN} & {(b)ENet-PAT} & {(c) TSRN-G} & {(d) TSRN-S} & {(e) Original } \\
	\end{tabular}
	
	\caption{Visual Comparison of recent state-of-the-art methods based on distortion metrics and perceptual quality with our texture based 4$\times$ image super-resolution.}
	\label{fig:first}
\end{figure}

\begin{figure}[t]
	\centering
	
	\begin{tabular}{c@{\hspace{0.01\linewidth}}c@{\hspace{0.01\linewidth}}c@{\hspace{0.01\linewidth}}c@{\hspace{0.01\linewidth}}c@{\hspace{0.01\linewidth}}c@{\hspace{0.01\linewidth}}c@{\hspace{0.01\linewidth}}c}
		
		\includegraphics[width = .13\linewidth]{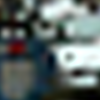} &
		\includegraphics[width = .13\linewidth]{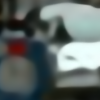} &
		\includegraphics[width = .13\linewidth]{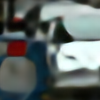} &
		\includegraphics[width = .13\linewidth]{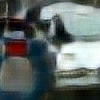} &
		\includegraphics[width = .13\linewidth]{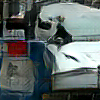} &
		\includegraphics[width = .13\linewidth]{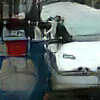} &
		\includegraphics[width = .13\linewidth]{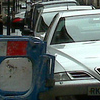} \\
		
		\includegraphics[width = .13\linewidth]{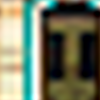} &
		\includegraphics[width = .13\linewidth]{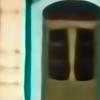} &	
		\includegraphics[width = .13\linewidth]{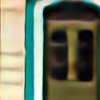} &
		\includegraphics[width = .13\linewidth]{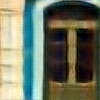} &
		\includegraphics[width = .13\linewidth]{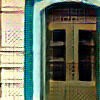} &
		\includegraphics[width = .13\linewidth]{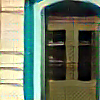} &
		\includegraphics[width = .13\linewidth]{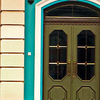} \\
		
		\includegraphics[width = .13\linewidth]{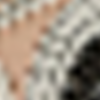} &
		\includegraphics[width = .13\linewidth]{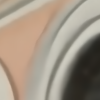} &
		\includegraphics[width = .13\linewidth]{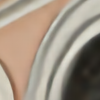} &
		\includegraphics[width = .13\linewidth]{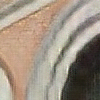} &
		\includegraphics[width = .13\linewidth]{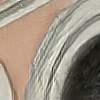} &
		\includegraphics[width = .13\linewidth]{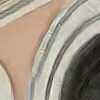} &
		\includegraphics[width = .13\linewidth]{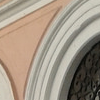} \\
		
		{(a)Bicubic } & {(b)LapSRN }& {(c)SRresnet} & {(d)SRGAN} & {(e)TSRN-G} & {(f)TSRN-S} & {(g)Original } \\
	\end{tabular}
	
	\caption{Visual Comparison of recent state-of-the-art methods based on distortion metrics and perceptual quality with our texture based  8$\times$ image super-resolution.}
	\label{fig:fourth}
\end{figure}

In Fig.~\ref{fig:first} and \ref{fig:fourth} we show visual comparisons with recently proposed state-of-the-art models for both 4$\times$ and 8$\times$ super-resolution. Our TSRN model manages to hallucinate realistic textures and image details and compares favorably with the state-of-the-art.

\subsection{TSRN-Faces on CelebA Dataset.}
In addition to training on MS-COCO dataset\cite{lin2014microsoft}, we also tested our proposed texture based super resolution method for CelebA faces dataset\cite{liu2015faceattributes}. Our method yields visible improvements over other methods. More specifically we compare with Enhancenet-PAT \cite{SajSchHir17} which employs GAN for enhancing textures. We observe that such method has a tendency to manipulate the overall facial features, thus not maintaining the integrity of the input image. In comparison, our method learns the texture mapping between a low resolution image ($I_{LR}$) and its high resolution counterpart ($I_{HR}$) thus generates visually plausible results.

\section{Using Texture as a Perceptual Metric}
\begin{figure}[h]
	\centering
	\begin{tabular}{c@{\hspace{0.01\linewidth}}c@{\hspace{0.01\linewidth}}c}
		\includegraphics[width = .505\linewidth]{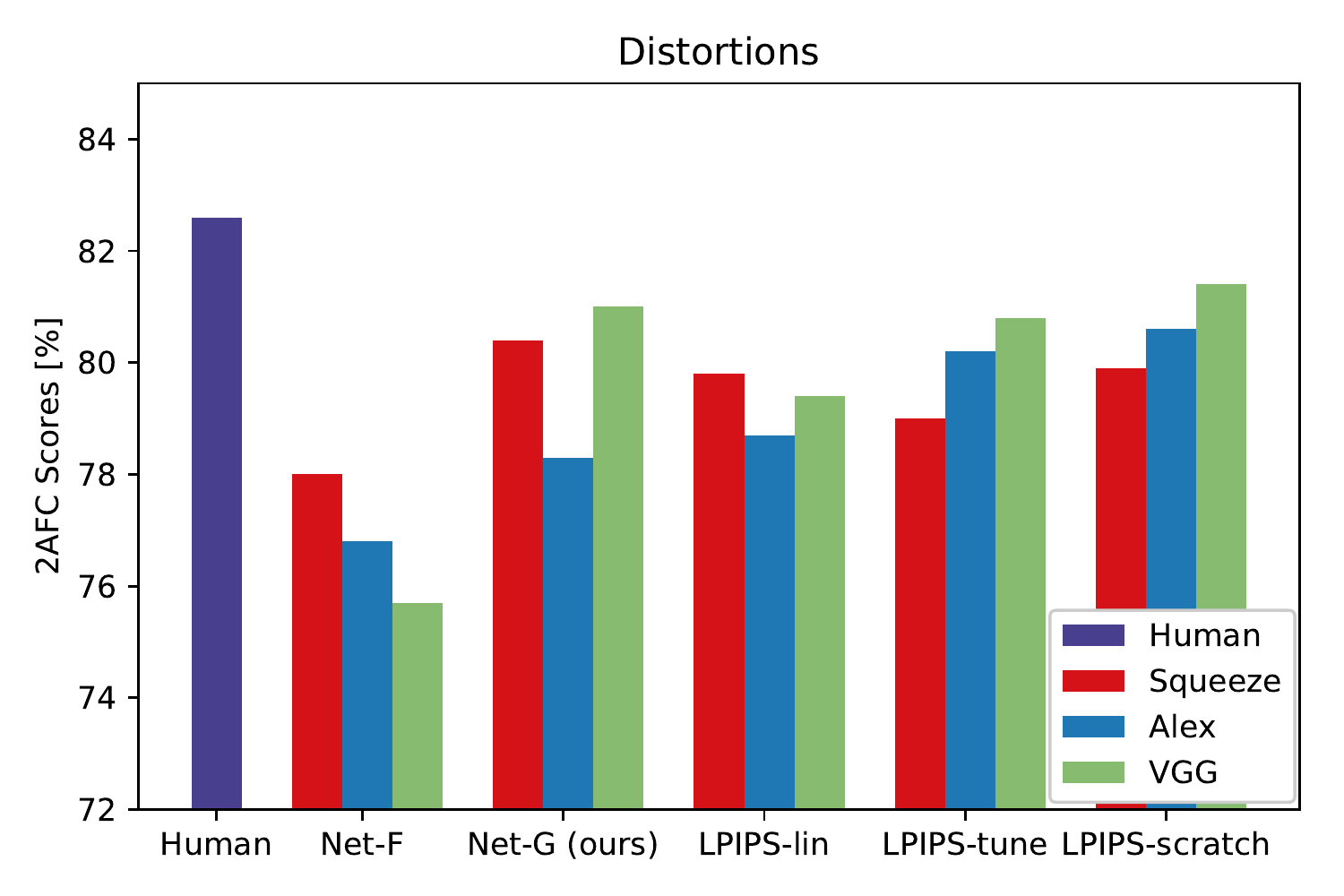} &		
		\includegraphics[width = .505\linewidth]{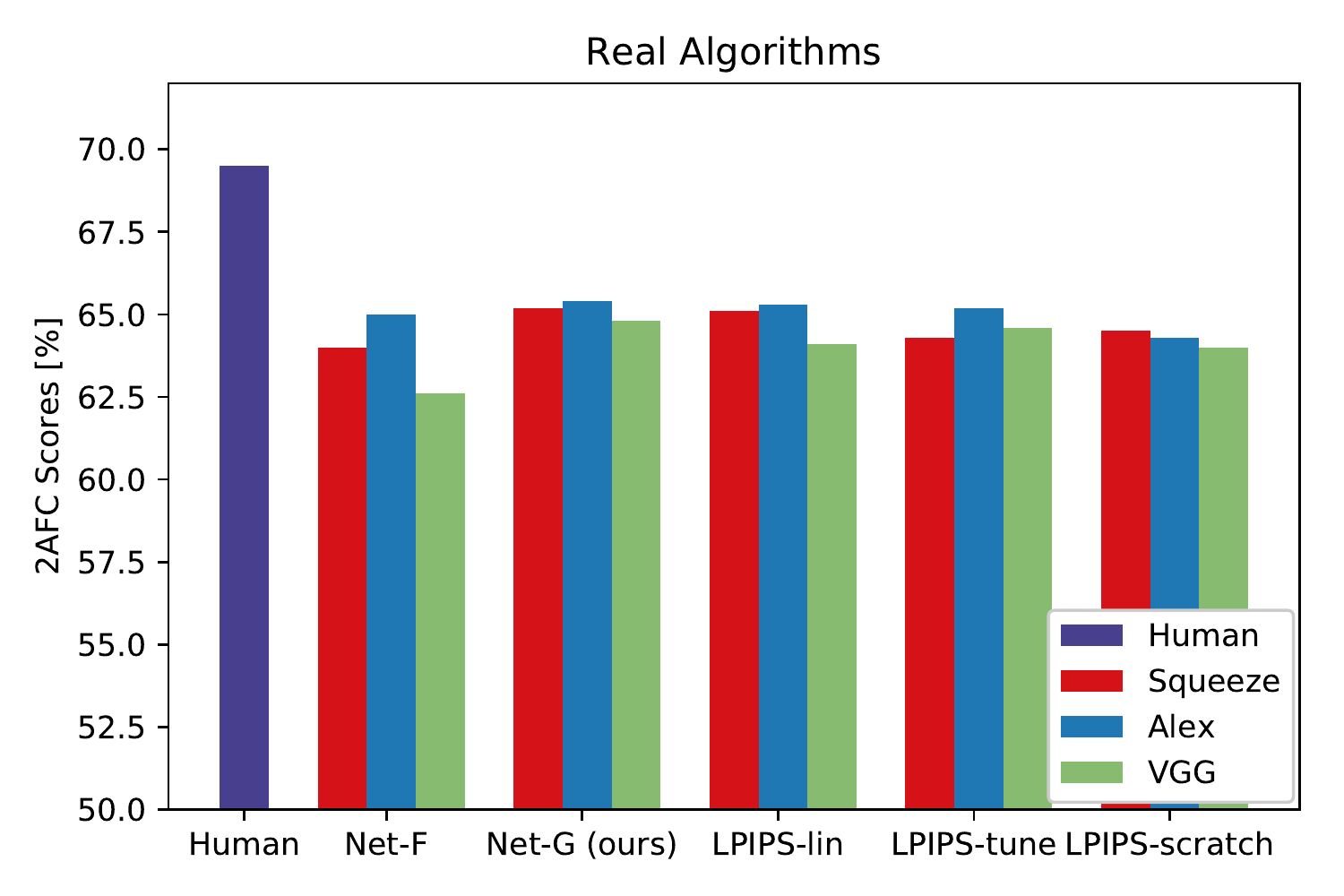} 
	\end{tabular}
	\caption{Quantitative comparison between different methods for determining perceptual similarity on the BAPPS validation dataset~\cite{zhang2018unreasonable}. Our Gram matrices based distance (Net-G) scores better than the feature based method (Net-F). Net-G results are comparable to calibrated *LPIPS metrics which are specifically trained on BAPPS training dataset, thus have an advantage.}
	\label{fig:lpips}
\end{figure}

In this section, we propose an improvement on LPIPS \cite{zhang2018unreasonable}, a recently proposed perceptual similarity metric based on deep features. The method computes the distance between the deep features of two images in order to determine the perceptual similarity between them. We argue that Gram matrices that measure the correlations of the same deep features, provide a richer and better framework for capturing the perceptual representation of images than the features themselves. Therefore, instead of computing the distances between the features of a given convolutional layer, we compute the distance between their Gram matrices. For a pair of reference and distorted patches (\textit{x,$x_0$}), we compute their normalized Gram matrices $\hat{G}^l$ and $\hat{A}^l$ $ \in \mathbb{R}^{C_l \times C_l}$, where $C$ is the number of channels in layer $l$. We compute the distance between them using the same formulation as in equation\ref{eqtn:loss} and then sum it up across all layers $l$, i.e.
\begin{equation}
d(x,x_0) = \sum_{l} \frac{1}{C_l^2} \sum_{i=1}^{C_l} \sum_{j=1}^{C_l} (G_{i,j}^l - A_{i,j}^l)^2
\label{eqtn:lpips}
\end{equation}
Using the features of ``uncalibrated" pre-trained image classification networks, this Gram matrices distance achieves better 2AFC scores on the BAPPS validation dataset than the distances based on the features themselves. As shown in \figurename{\ref{fig:lpips}}, our results (Net-G) are comparable to the ``calibrated" LPIPS models (specifically trained on BAPPS training datasets) and also outperform them in some benchmarks. For comparison, we adopted the same configuration of three reference models (SqueezeNet\cite{iandola2016squeezenet}, AlexNet\cite{krizhevsky2012imagenet} and VGG-16\cite{simonyan2014very}) used by \cite{zhang2018unreasonable}. However, to get the best results we changed the number of layers for the distance computation, more specifically we did not use the feature activations before the first pooling layer and after the penultimate pooling layer of each model. This is because the texture from the lowest layers do not contain any structure in them whereas the last layers capture abstract and semantically more meaningful representation of the image but they lack in their ability to capture the perceptual details \cite{gatys2015texture}.
\begin{figure}[h]
	\centering
	
	\begin{tabular}{c@{\hspace{0.01\linewidth}}c@{\hspace{0.01\linewidth}}c@{\hspace{0.01\linewidth}}c@{\hspace{0.01\linewidth}}c@{\hspace{0.01\linewidth}}c}
		
		\includegraphics[width = .185\linewidth]{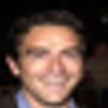} &
		\includegraphics[width = .185\linewidth]{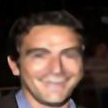} &
		\includegraphics[width = .185\linewidth]{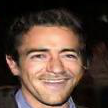} &
		\includegraphics[width = .185\linewidth]{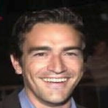} &
		\includegraphics[width = .185\linewidth]{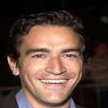} \\
		
		\includegraphics[width = .185\linewidth]{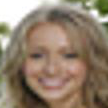} &
		\includegraphics[width = .185\linewidth]{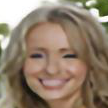} &
		\includegraphics[width = .185\linewidth]{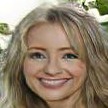} &
		\includegraphics[width = .185\linewidth]{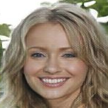} &
		\includegraphics[width = .185\linewidth]{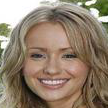} \\

		\includegraphics[width = .185\linewidth]{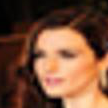} &
		\includegraphics[width = .185\linewidth]{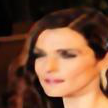} &
		\includegraphics[width = .185\linewidth]{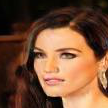} &
		\includegraphics[width = .185\linewidth]{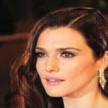} &
		\includegraphics[width = .185\linewidth]{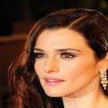} \\
		
		{(a) Bicubic }& {(b) ENet-PAT} & {(b) ENet-PAT-F} & {(c) TSRN-Faces} & {(d) Original} \\
	\end{tabular}
	
	\caption{Visual comparison of different networks trained on CelebA dataset \cite{liu2015faceattributes} for 4$\times$ SISR. TSRN yields visually faithful results to the original input image.}
	\label{celeba}
\end{figure}

\begin{table*}[t]
	\begin{center}
		\resizebox{\linewidth}{!} {
			\begin{tabular}{l l@{\hskip .05in} c@{\hskip .05in} c@{\hskip .05in} c@{\hskip .15in} c@{\hskip .05in} @{\hskip .05in}c@{\hskip .05in} c@{\hskip .05in} c@{\hskip .05in} c@{\hskip .15in} c@{\hskip .05in} } \toprule
				
				\multirow{3}[3]{*}{\textbf{Subtype}} &\multirow{3}[3]{*}{\textbf{Metric}} & \multicolumn{3}{c}{\textbf{Distortions}} & \multicolumn{5}{c}{\textbf{Real Algorithms}} & \multirow{1}{*}{\textbf{All}} \\ \cmidrule(lr){3-5} \cmidrule(lr){6-10} \cmidrule(lr){11-11} 
				
				& & \textbf{Trad-} & \textbf{CNN-} & \multirow{2}{*}{\textbf{All}} & \textbf{Super-} & \textbf{Video} & \textbf{Color-} & \textbf{Frame} & \multirow{2}{*}{\textbf{All}} & \multirow{2}{*}{\textbf{All}} \\
				
				& & \textbf{itional} & \textbf{Based} & \textbf{} & \textbf{res} & \textbf{Deblur} & \textbf{ization} & \textbf{Interp} & \\ \midrule
				
				Oracle & Human & 80.8 & 84.4 & 82.6 & 73.4 & 67.1 & 68.8 & 68.6 & 69.5 & 73.9 \\ \midrule
				
				\multirow{7}{*}{} & Squeeze -- lin & 76.1 & \textcolor{red}{\textbf{83.5}} & 79.8 & 71.1 & 60.8 & \textcolor{red}{\textbf{65.3}} & 63.2 & 65.1 & \textcolor{red}{\textbf{70.0}} \\
				\multirow{7}{*}{*LPIPS \cite{zhang2018unreasonable}} & Alex -- lin & 73.9 & 83.4 & 78.7 & 71.5 & \textcolor{blue}{\textbf{61.2}} & \textcolor{red}{\textbf{65.3}} & 63.2 & \textcolor{red}{\textbf{65.3}} & 69.8 \\
				\multirow{7}{*}{} & VGG -- lin & 76.0 & 82.8 & 79.4 & 70.5 & 60.5 & 62.5 & 63.0 & 64.1 & 69.2 \\ \cdashline{2-11}
				& Squeeze -- scratch & 74.9 & 83.1 & 79.0 & 71.1 & 60.8 & 63.0 & 62.4 & 64.3 & 69.2 \\
				& Alex -- scratch & 77.6 & 82.8 & 80.2 & 71.1 & 61.0 & \textcolor{blue}{\textbf{65.6}} & \textcolor{red}{\textbf{63.3}} & 65.2 & \textcolor{blue}{\textbf{70.2}} \\
				& VGG -- scratch & 77.9 & \textcolor{blue}{\textbf{83.7}} & 80.8 & 71.1 & 60.6 & 64.0 & 62.9 & 64.6 & 70.0 \\ \cdashline{2-11}
				& Squeeze -- tune & 76.7 & 83.2 & 79.9 & 70.4 & \textcolor{red}{\textbf{61.1}} & 63.2 & 63.2 & 64.5 & 69.6 \\
				& Alex -- tune & 77.7 & 83.5 & 80.6 & 69.1 & 60.5 & 64.8 & 62.9 & 64.3 & 69.7 \\
				& VGG -- tune & \textcolor{blue}{\textbf{79.3}} & \textcolor{red}{\textbf{83.5}} & \textcolor{blue}{\textbf{81.4}} & 69.8 & 60.5 & 63.4 & 62.3 & 64.0 & 69.8 \\ \midrule
				\multirow{3}{*}{Supervised-} & SqueezeNet~\cite{iandola2016squeezenet}& 73.3 & 82.6 & 78.0 & 70.1 & 60.1 & 63.6 & 62.0 & 64.0 & 68.6 \\
				\multirow{3}{*}{Nets \cite{zhang2018unreasonable}}& AlexNet~\cite{krizhevsky2012imagenet}& 70.6 & 83.1 & 76.8 & \textcolor{blue}{\textbf{71.7}} & 60.7 & 65.0 & 62.7 & 65.0 & 68.9 \\
				& VGG~\cite{simonyan2014very}& 70.1 & 81.3 & 75.7 & 69.0 & 59.0 & 60.2 & 62.1 & 62.6 & 67.0 \\ 	
				\midrule
				\multirow{3}{*}{Supervised-} & SqueezeNet~\cite{iandola2016squeezenet} & 77.5 & 83.2 & 80.4 & \textcolor{red}{\textbf{71.6}} & \textcolor{red}{\textbf{61.1}} & 65.1 & 62.9 & 65.2 & \textcolor{blue}{\textbf{70.2}} \\
				\multirow{3}{*}{Nets (Ours)} & AlexNet~\cite{krizhevsky2012imagenet}& 73.5 & 83.0 & 78.3 & 71.5 & 60.9 & \textcolor{blue}{\textbf{65.6}} & \textcolor{blue}{\textbf{63.4}} & \textcolor{blue}{\textbf{65.4}} & 69.7 \\
				& VGG~\cite{simonyan2014very}& \textcolor{red}{\textbf{78.3}} & \textcolor{blue}{\textbf{83.7}} & \textcolor{red}{\textbf{81.0}} & 70.9 & 60.9 & 64.3 & 63.1 & 64.8 & \textcolor{blue}{\textbf{70.2}} \\
				\bottomrule
			\end{tabular}
		}
		\caption{ 2AFC scores (higher is better) for different methods using disparity in deep feature representations \cite{zhang2018unreasonable} and texture representations (ours) on BAPPS validation dataset. Values in blue are \textcolor{blue}{highest performing} while the values in red are \textcolor{red}{the second best}. Our texture based scores from untrained supervised networks consistently perform better than the feature based scores and compare to *LPIPS metrics which are specifically trained on BAPPS training dataset, thus have an advantage over other untrained methods.}
		\label{tab:res_quant}
	\end{center}
\end{table*}
\section{Conclusion}
Transferring texture via matching Gram matrices has been very successful in image style transfer, however their utility for natural image enhancement has not been studied extensively. In this work we demonstrate that Gram matrices are very powerful in capturing perceptual representations of images which makes them a perfect candidate for their use in a perceptual similarity metric like LPIPS. 
Exploiting this ability, we obtain image reconstructions of high perceptual quality for the task of 4$\times$ and 8$\times$ single image super-resolution. We further devise a scheme for external semantic guidance for controlling texture transfer which is particularly helpful for 8$\times$ super-resolution. Our method is simple, easily reproducible and yet effective. We believe that texture loss can have far reaching implications in the future research of image restoration.

\clearpage
{\small
\bibliographystyle{splncs}
\bibliography{egbib}}

\section{Appendix}
\subsection{Constraining Texture Transfer.}
In this section we try to understand how different textures get hallucinated and how important is the role of content image's features in determining the mapping of textures from a style image.

It is necessary to understand the role of pre-trained VGG-19's \cite{simonyan2014very} convolutional kernels which form the basis of Gram matrices composition. The kernels are translational invariant as they are able to look out for the specific features (for example, an oriented edge in an image) using the same weights after sweeping across the entire image. A specific feature in the entire image space would always be detected by a specific kernel \cite{dieleman2015rotation}. This means that if both $I_{s}$ and $I_{c}$ have identical features then they are always going to be detected by the same kernels. Therefore, the activations in resulting feature maps $F^l \in \mathbb{R}^{N_l \times M_l}$ and $P^l \in \mathbb{R}^{N_l \times M_l}$, where $N_l$ is the number of feature maps in layer \textit{l} and $M_l$ is the product of height and width of feature maps in layer \textit{l} i.e. $M_l =  height \times width$, would correspond to each other in channel dimensions $N_l$ (each kernel results in one feature map). However, the kernels are not rotational and scale invariant thus different kernels would respond to a rotated version of the same feature and the correspondence amongst feature maps' activations in $N_l$ dimension gets lost.\\

Gram matrices $G^l$ and $A^l$ capture the spatial correlations of feature maps $F^l$ and $P^l $, the presence of identical features inside $I_{s}$ and $I_{c}$ results in the correspondences of Gram matrices activations as well.
\begin{figure}[h!] 
	\centering
	\includegraphics[width=1.0\linewidth]{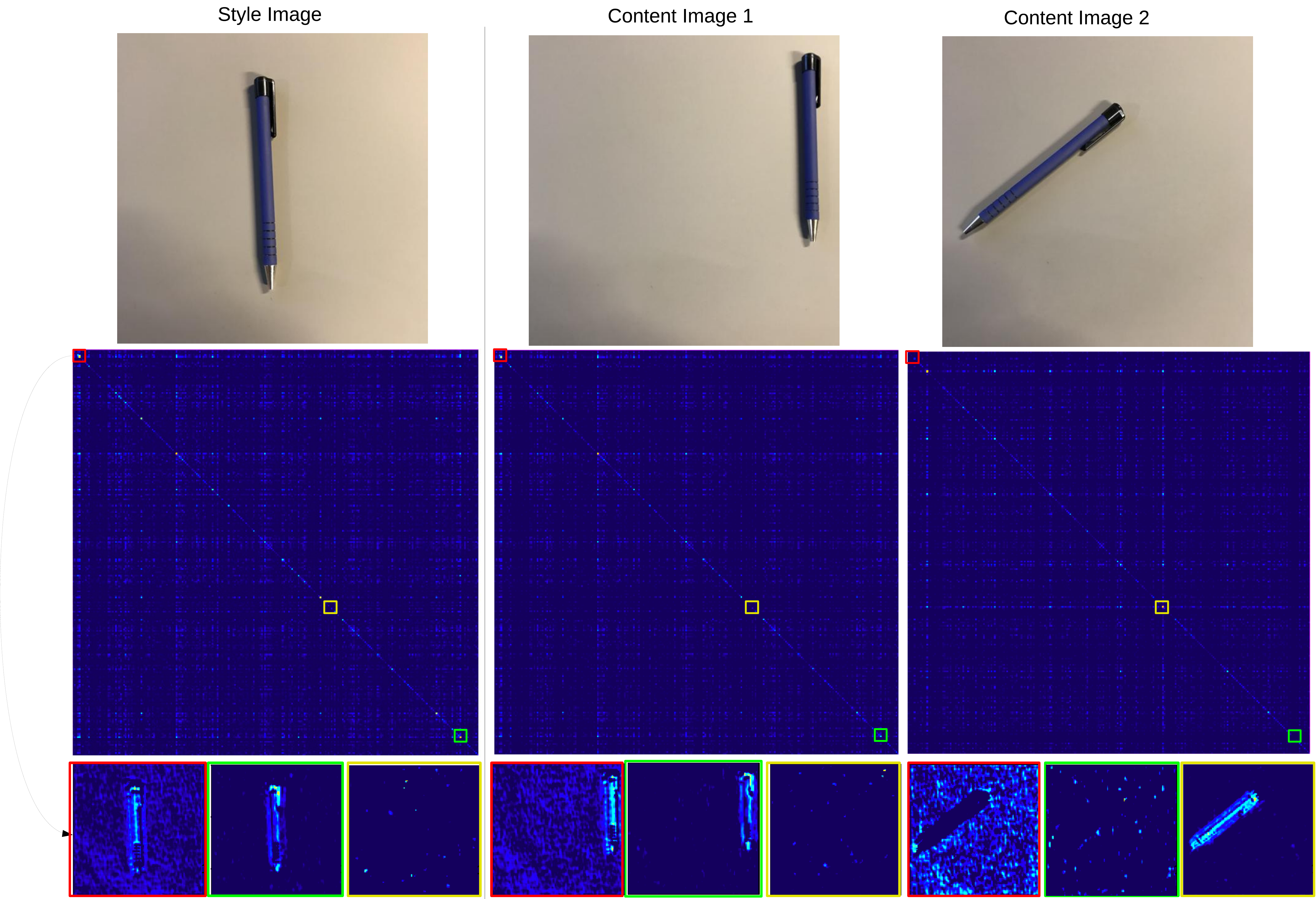}\hspace{0.1em}
	\vspace{-1.5em}
	\caption{The top row shows the style and content images, whereas their corresponding Gram matrices are shown in the middle. The bottom row shows the feature activation maps corresponding to the peak activations in the Gram matrices. The feature maps in red and green boxes correspond to the top 2 activations in style image Gram matrix and the respective feature maps of both content images at those locations. The yellow box corresponds to the peak activation in content image 2's Gram matrix. View electronically for better analysis. }
	\label{fig:exp2}
\end{figure}
Similarly, if both $I_{s}$ and $I_{c}$ do not possess identical features or the features exhibit rotational variation then the activation pattern in Gram matrices changes. In \figurename{\ref{fig:exp2}}, we present a simple real world example of natural images which shows how the variations in global scenes affect the correspondence between the activations of $F^l$ and $P^l$ and thus the activation pattern of Gram matrices $G^l$ and $A^l$. We visualize the Gram matrices of the feature maps from the layer \textit{conv3\_4} of VGG-19. We specifically chose this intermediate layer as it is sensitive enough to intermediate features (neither too abstract nor too low level).  We can see that the activation pattern of the Gram matrices corresponding to the style image and the content image 1 look very similar, even though the content image shows the variation in translation. Whereas, the activation pattern of the Gram matrix corresponding to content image 2 looks completely different even though the image has the same content as that of the style image. Such phenomenon is inevitable since the convolutional kernels are translational invariant and not rotationally invariant. Thus a small change in orientation of the pen instigates different kernels.\\

Since each entry of the Gram matrix is a summed up value of spatially correlated feature maps, therefore for a more detailed analysis we look into the feature maps corresponding to the peak activations of style image's Gram matrix, shown in red and green boxes. We also show the corresponding feature maps of both content images to which they get mapped to during optimization. The feature maps of the style image and the content image 1 show huge correlations in their activations. On the other hand, we see very weak correlations in the corresponding feature map activations of the style image and the content image 2. The yellow box corresponds to the peak activation in content image 2's Gram matrix. The corresponding feature map (in yellow box) of the style image, to which it gets mapped to, contains very less information.

\subsubsection{Mapping of Features in Gram Matrices}

While optimizing for $\mathcal{L}_{texture}$ in equation \ref{eqtn:loss}, we reduced the mean squared error between these Gram matrices values. Each entry of the Gram matrix represents the summed up statistics of feature maps' correlations, where we have already lost the spatial information. Thus the optimization only enforces the summed up statistics on each entry of Gram matrix and not the spatial constraints. 
In \figurename{\ref{fig:exp2}}, using the Gram matrix structure, different feature maps correlations of $I_{s}$ and $I_{c}$ get mapped to each other. We observe that this mapping plays a huge role in the hallucination of textures in the content images.

\begin{figure}[h!] 
	\centering
	\includegraphics[width=1.0\linewidth]{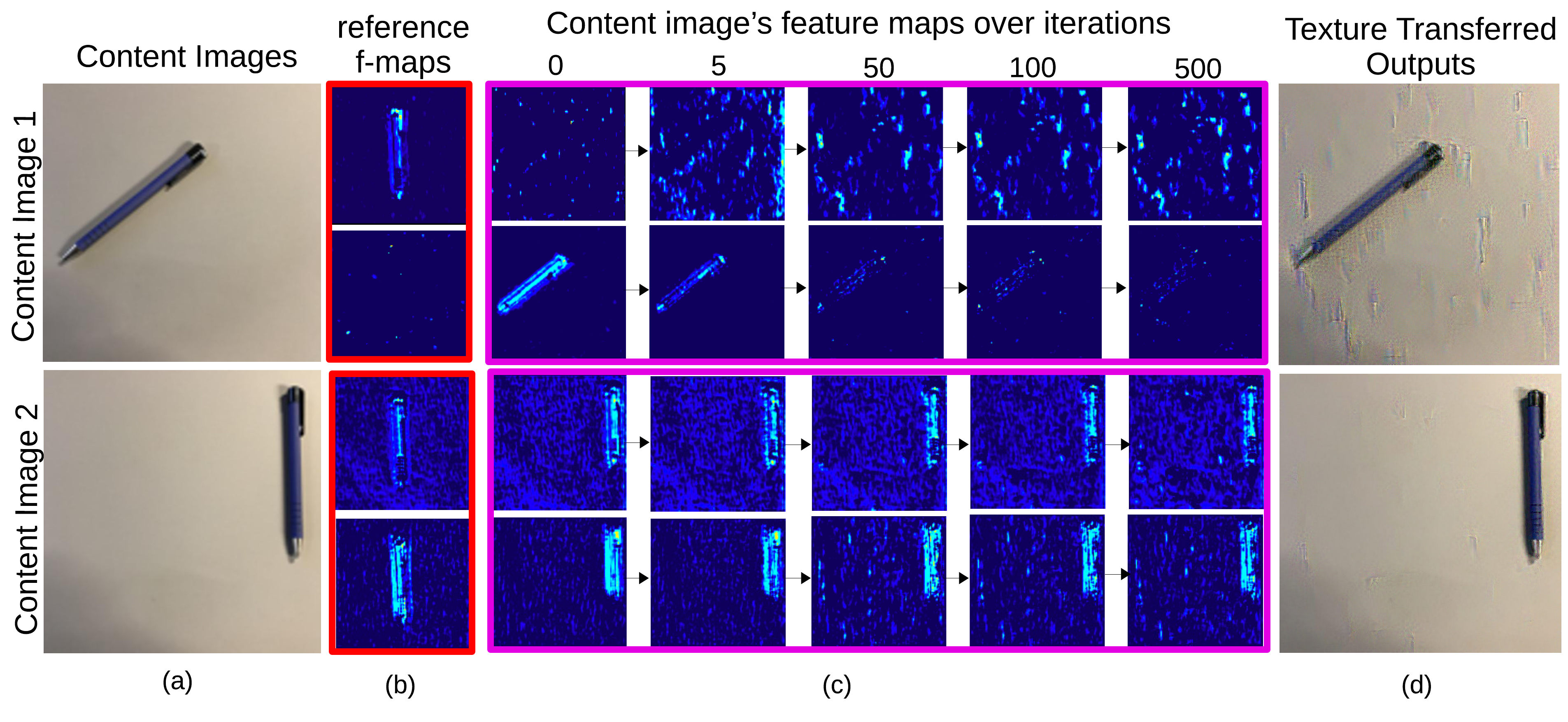}\hspace{0.1em}
	\vspace{-1.5em}
	\caption{The left image (a) shows the initial content images which are first $4\times$ bi-cubically downsampled and then upsampled back (b) shows the reference feature maps of style image from \figurename{\ref{fig:exp2}}, the top image refers to the location of peak activation in style image's Gram matrix and the bottom one refers to the content image's. The middle figure (c) shows the evolution of respective feature maps of the content images. (d) shows the resultant image.}
	\label{fig:exp3}
\end{figure}

\figurename{\ref{fig:exp3}} shows how the feature representation of the reference style image in \figurename{\ref{fig:exp2}} are enforced on the content image's feature maps during optimization. For the content image 1, there is huge mismatch in the activation levels of the style image feature maps and the corresponding feature maps of the content image 1.
The top row shows how the information is enforced, while the bottom row shows how the information is diminished in the content image 1's features. As explained earlier, the Gram matrix entries are just the summed up feature correlations values, having no spatial information thus the texture get hallucinated randomly across the image. The VGG-19 kernels themselves remain unchanged during optimization, therefore an image search is performed in order to match the feature representations of the style image. This mismatch in the mappings of Gram matrix values explain how a small variation like rotation can induce big changes in the texture transfer. Similarly if we look at the evolution of feature maps of content image 2, then we see high correlation with the activation levels of the style image.

In style transfer \cite{gatys2016image}, the content and the style image have different features which result in the huge mismatch of their Gram matrices activation pattern, and consequently the feature maps mappings. This mismatch results in the hallucination of new textures.

\subsubsection{Exploiting Translational Invariant Mapping of Textures}

\begin{figure}[h!] 
	\centering
	\includegraphics[width=1.0\linewidth]{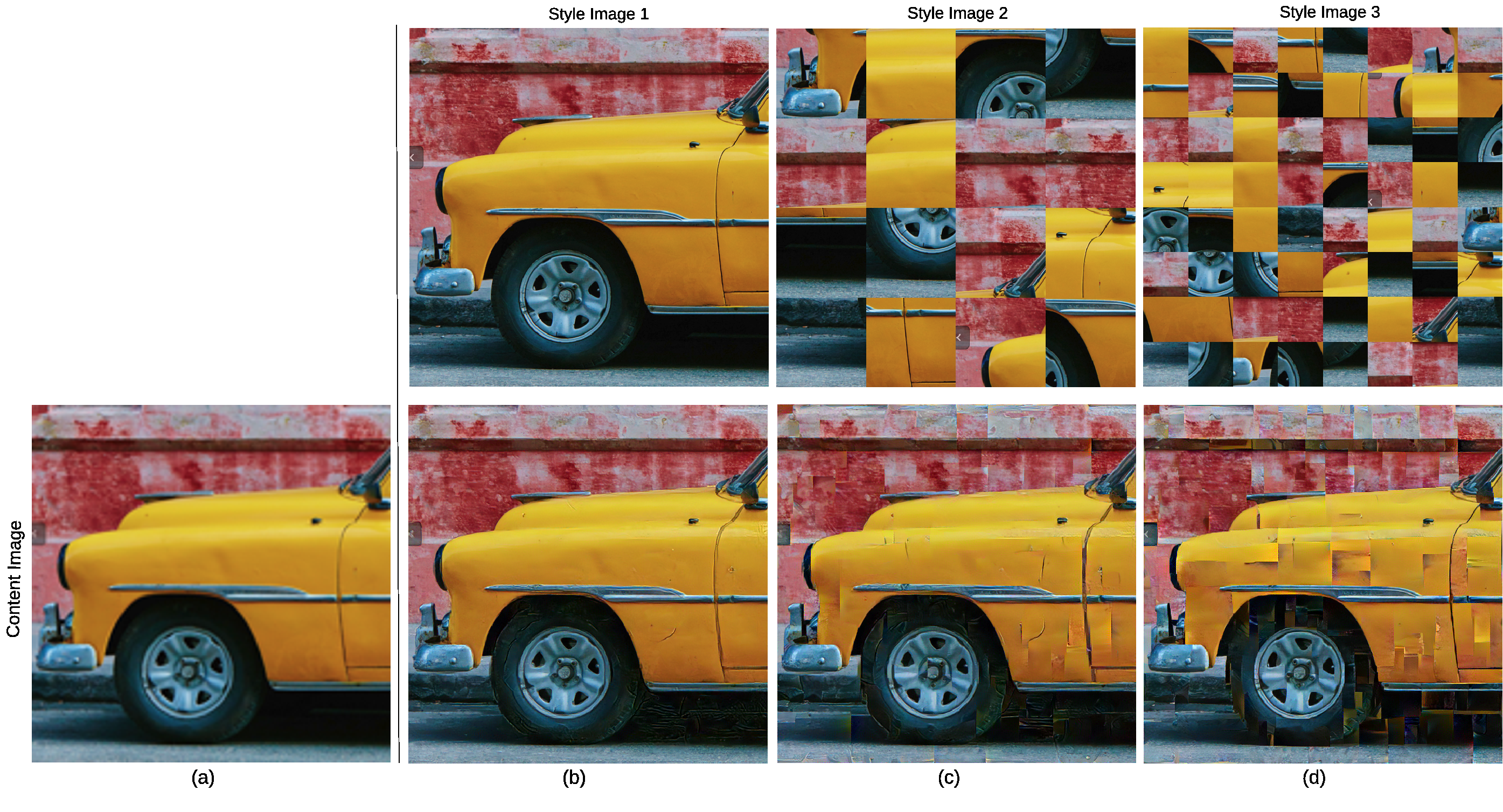}\hspace{0.1em}
	\vspace{-1.5em}
	\caption{The left image (a) shows the content image which is 4x bi-cubically downsampled and then up-sampled again bi-cubically. (b) shows the texture transfer from style image 1. (c) and (d) shows the texture transfer from style image 2 and 3. Notice that the style image 2 and 3 have randomly shuffled patches of $160\times160$ and $80\times80$ respectively. Even then the Gram matrices have been able to satisfactorily map the textures correctly. Zoom in for texture analysis.}
	\label{fig:exp4}
\end{figure}

We saw that the convolutional kernels extend their translational invariant property to the Gram matrices based texture transfer. In the following we discuss the efficiency and the implications of such texture mapping. In \figurename{\ref{fig:exp4}}(a), the content image is first bi-cubically down-sampled 4 times

 and then up-sampled back to the original dimensions bi-cubically. This is done to make sure that there is enough difference in the textures of the image pair which needs to be recovered via Gram matrices based optimization.

\figurename{\ref{fig:exp4}}(b) demonstrates the texture transfer from the style image 1. The existence of the identical features in both images enabled $\mathcal{L}_{t}$ to map the texture from the style image correctly. \figurename{\ref{fig:exp4}}(c) shows the texture transfer from a relatively complex style image which has randomly shuffled patches of size 80 x 80. Both content and style images have similar features (ignoring the new edge pattern in the style image). Now according to the generally known behavior of Gram matrices based texture transfer the texture from the style image should not follow any spatial constraint and get transferred to the content image in haphazard fashion. However, we observe that the VGG-19 convolutional kernels are unreasonably effective in capturing the identical features correctly and then Gram matrix composition provides a very convenient structure in their translational invariant mapping. Thus the transfer of texture respects the global spatial arrangement of the content image. We do observe some artifacts, mainly because of the new edge structure in the style image and also because of the existence of homogeneous regions in the content image where finding the relevant features for mapping is difficult.

\subsection{Model Agnostic Behavior.}
\begin{figure}[t]
	\centering
	
	\begin{tabular}{c@{\hspace{0.01\linewidth}}c@{\hspace{0.01\linewidth}}c@{\hspace{0.01\linewidth}}c@{\hspace{0.01\linewidth}}c@{\hspace{0.01\linewidth}}c}
		
		\includegraphics[width = .185\linewidth]{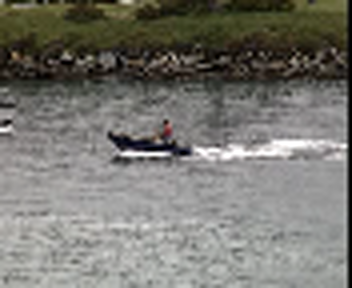} &
		\includegraphics[width = .185\linewidth]{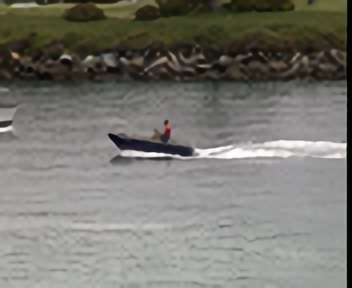} &
		\includegraphics[width = .185\linewidth]{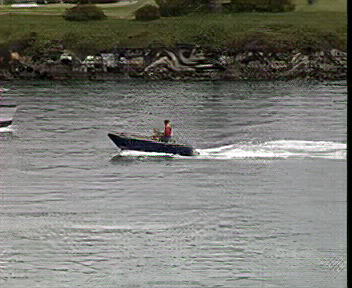} &
		\includegraphics[width = .185\linewidth]{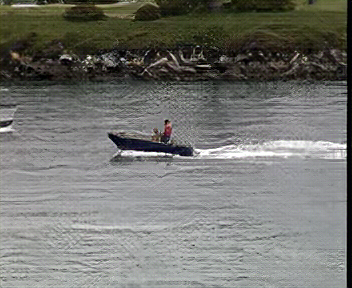} &
		\includegraphics[width = .185\linewidth]{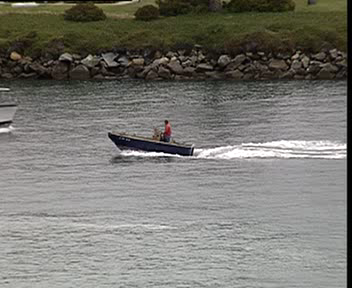} \\
		
		\includegraphics[width = .185\linewidth]{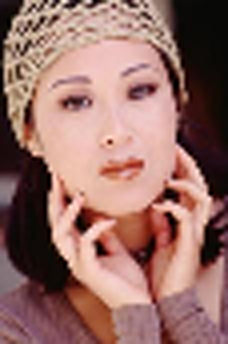} &
		\includegraphics[width = .185\linewidth]{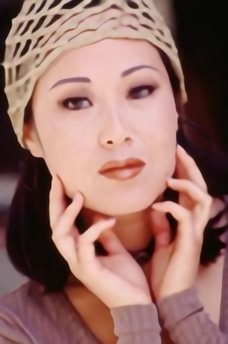} &
		\includegraphics[width = .185\linewidth]{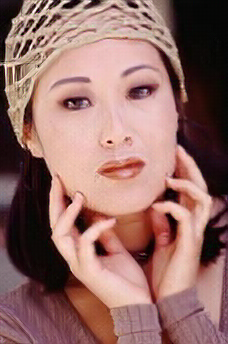} &
		\includegraphics[width = .185\linewidth]{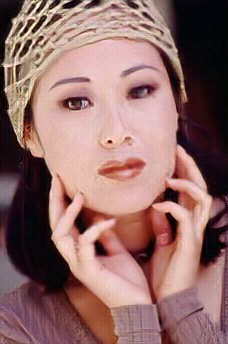} &
		\includegraphics[width = .185\linewidth]{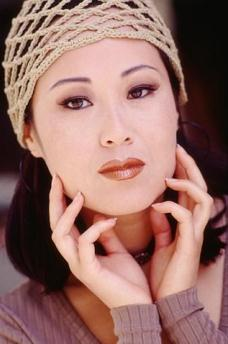} \\
		
		\includegraphics[width = .185\linewidth]{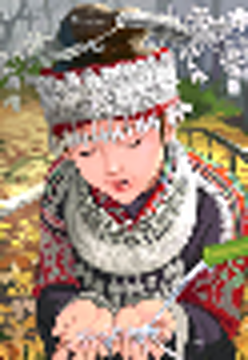} &
		\includegraphics[width = .185\linewidth]{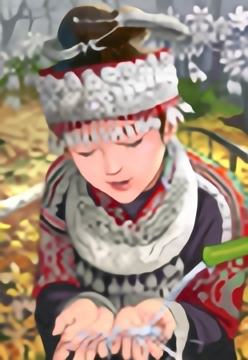} &
		\includegraphics[width = .185\linewidth]{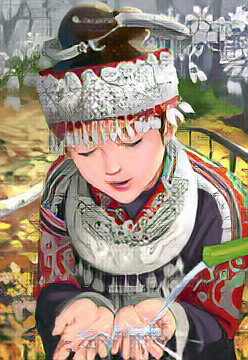} &
		\includegraphics[width = .185\linewidth]{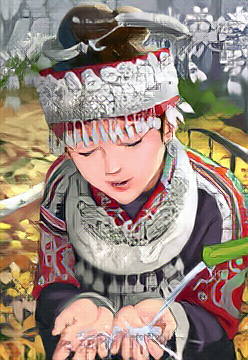} &
		\includegraphics[width = .185\linewidth]{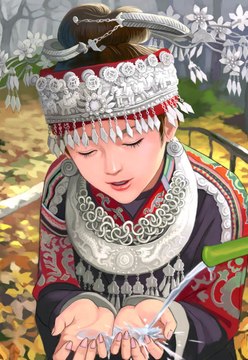} \\
		
		\includegraphics[width = .185\linewidth]{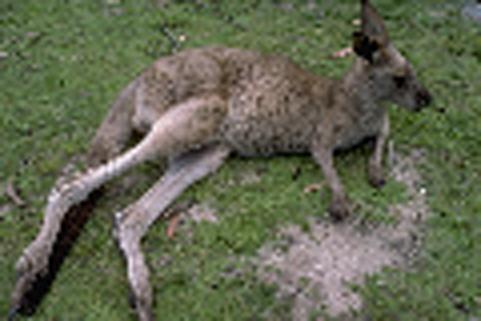} &
		\includegraphics[width = .185\linewidth]{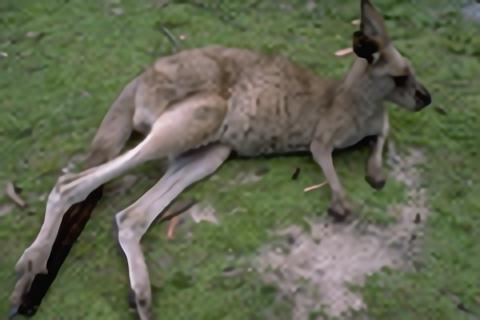} &
		\includegraphics[width = .185\linewidth]{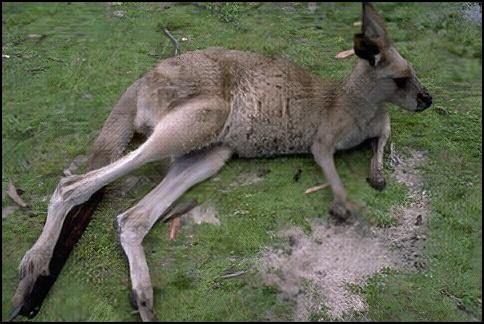} &
		\includegraphics[width = .185\linewidth]{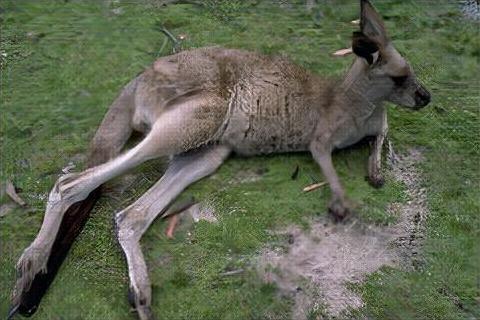} &
		\includegraphics[width = .185\linewidth]{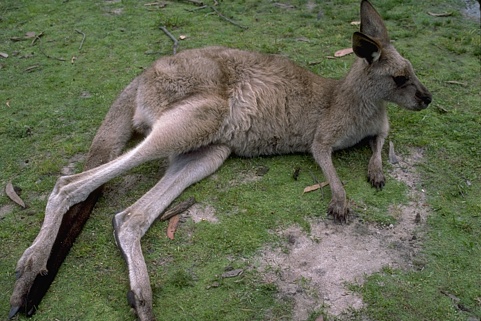} \\
		
		{(a) Bicubic }& {(b) MSLapSRN} & {(b) TSRN-G} & {(c) TSRN-Shallow} & {(d) Original} \\
	\end{tabular}
	
	\caption{Visual Comparison to analyze the model agnostic behavior of texture based 4$\times$ SISR. See that TSRN-Global and TSRN-small has very small visual differences. Zoom in for better analysis.}
	\label{shallow}
\end{figure}

Both TSRN-G and TSRN-S used relatively deep architectures for learning the mappings between $I_{LR}$ and $I_{HR}$. To test the effectiveness of texture based super resolution, we also trained with a relatively shallow model having 6 residual blocks and the number of convolutional kernels reduced from 64 to 32. This reduced model is named TSRN-shallow, where the network's parameters are drastically reduced from $\approx$1.07M to $\approx$195k and the required memory storage from 4.3MB to 832.7 kB. Surprisingly, we did not observe much qualitative differences in the results, shown in \figurename{ \ref{shallow}}. However, the LPIPS metric evaluated TSRN-S and TSRN-G to be superior than TSRN-shallow, shown in Table~\ref{tab:comp_1}. TSRN-shallow is trained the same way, that is pre-training with MSE for 10 epochs and then training with texture loss for another 100 epochs.
\begin{table*}[h!]
	\begin{center}
		\resizebox{\linewidth}{!} {
			\begin{tabular}{l@{\hskip .05in} c@{\hskip .05in} c@{\hskip .05in} c@{\hskip .15in} c@{\hskip .05in} @{\hskip .05in}c@{\hskip .05in} c@{\hskip .05in} c@{\hskip .05in} c@{\hskip .15in}} \toprule
				\centering
				\multirow{2}[3]{*}{\textbf{Metric}} & \multicolumn{2}{c}{\textbf{Set 5}} & \multicolumn{2}{c}{\textbf{Set 14}} &
				\multicolumn{2}{c}{\textbf{BSD 100}} &
				\multicolumn{2}{c}{\textbf{Urban}}\\ 
				\cmidrule(lr){2-3} \cmidrule(lr){4-5} \cmidrule(lr){6-7} \cmidrule(lr){8-9}
				\centering
				& \textbf{AlexNet} & \textbf{SNet} & \textbf{AlexNet} & \textbf{SNet}  & \textbf{AlexNet} & \textbf{SNet} & \textbf{AlexNet} & \textbf{SNet} \\ \midrule
				
				Bicubic & 0.1585 & 0.1202 & 0.1731& 0.1320 & 0.1463 &0.1007 & 0.1552& 0.1238 \\
				TSRN-S &0.0273&0.0394&0.0438& 0.0483& \textbf{0.0478}&0.0420&0.0397&0.0404\\  
				TSRN-G & 0.0285&\textbf{0.0358} & 0.0463& \textbf{0.0456} & 0.0481 &\textbf{0.0404} & \textbf{0.0385}& \textbf{0.0392}\\
				\midrule
				TSRN-shallow & 0.0287&0.0397 & \textbf{0.0388} &0.0473 & 0.0483&0.0406 & 0.0431& 0.0415\\				
				\bottomrule
			\end{tabular}
		}
		\vspace{1mm}
		\caption{Comparison for 4$\times$ SISR on pre-trained AlexNet-linear and SqueezeNet-linear LPIPS metric\cite{zhang2018unreasonable}. Lower score is better.}
		\label{tab:comp_1}
		\vspace{-8mm}
	\end{center}
\end{table*}

\subsection{Semantically Controlled Fast Style Transfer.}
\begin{figure}[h!]
	\includegraphics[width=\linewidth]{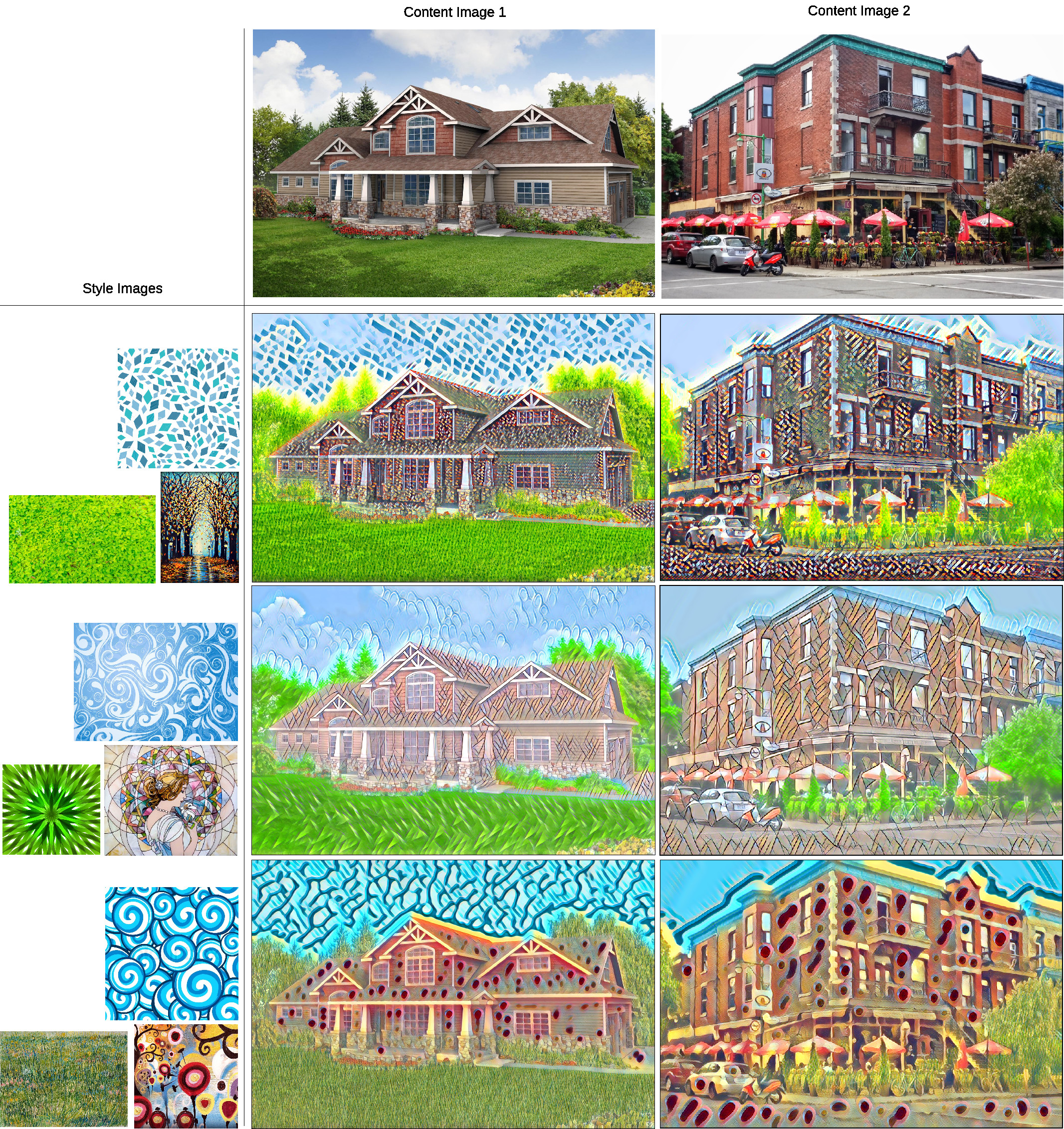}
	\caption{To show the effectiveness of our semantically controlled fast style transfer scheme. The images on the left show multiple style images which are semantically constrained to the sky, green and ``others" regions of the content images 1 and 2 respectively.}
	\label{semantic_transfer}
\end{figure}

In this section, we demonstrate the effectiveness of our semantically controlled texture transfer scheme in training a convolutional neural network architecture. We use the same architecture and loss formulation as provided by Johnson et al. \cite{johnson2016perceptual}. To train the networks for spatially controlled texture transfer, we used ADE20k dataset \cite{zhou2016semantic}. We made semantic division of the images into three classes i.e. sky, grass and others. Using the semantic masks provided with the dataset, we combined the classes of trees, fields, grass and plants into the ``grass" class. The sky class remained the same as provided with the dataset whereas the rest of the classes were grouped in ``others" class. Using the same methodology of semantic style transfer as explained in the main paper, we enforce different style images for these segmented regions. \figurename{ \ref{semantic_transfer}} depicts the effectiveness of our spatially controlled fast style transfer scheme for multiple style images.

\subsection{Architecture.}
\begin{figure}[h!]
	\includegraphics[width = .985\linewidth]{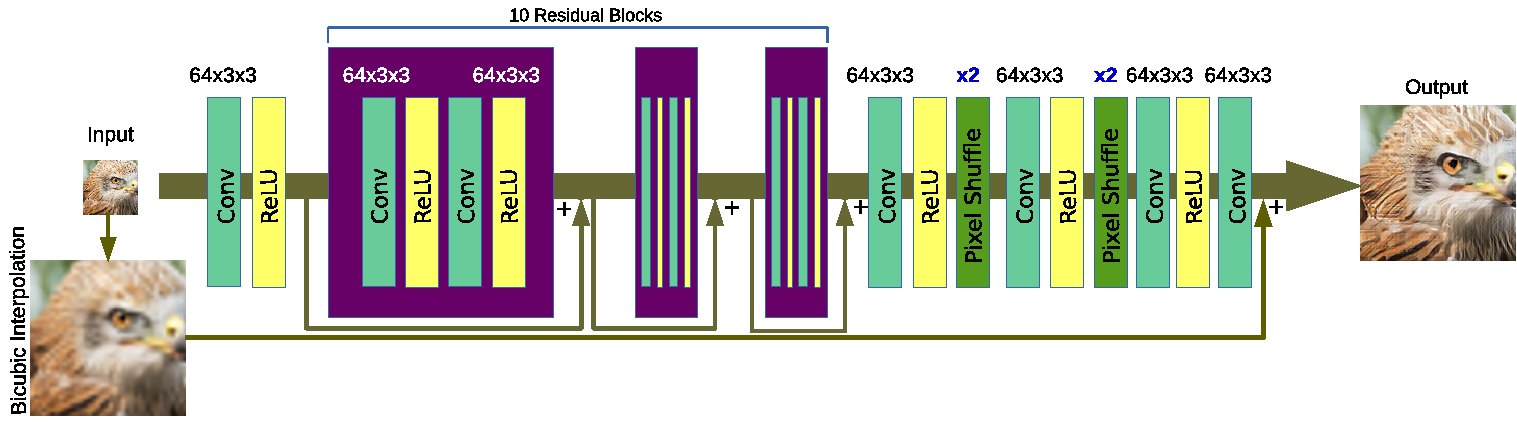}
	\caption{Pictorial illustration of the TSRN architecture used in the paper.}
	\label{architecture}
\end{figure}

The architecture for texture based super resolution network (TSRN) is inspired by \cite{SajSchHir17}. Similar form of architectures are used by many entries of NTIRE 2017 competition\cite{timofte2017ntire} and also by \cite{ledig2016photo}. The pictorial depiction of the architecture is shown in \figurename{ \ref{architecture}}. It must be noted that the architectures for the implementation of both global (TSRN-G) and semantically (TSRN-S) transferred texture methods has 10 residual blocks in each. This depth in the architecture is found to be enough to capture the texture mapping between a low resolution image $I_{LR}$ and its high resolution counterpart $I_{HR}$. 

\end{document}